\documentclass[runningheads]{llncs}

\usepackage{eccv}
\usepackage{eccvabbrv}
\usepackage{graphicx}
\usepackage{booktabs}
\usepackage[accsupp]{axessibility}  

\usepackage[pagebackref]{hyperref}

\usepackage{orcidlink}
\usepackage{graphicx}
\usepackage{latexsym}
\makeatletter
\@ifpackageloaded{hyperref}{
}{
  \usepackage[pagebackref,breaklinks,colorlinks]{hyperref}
}
\@ifpackageloaded{prettyref}{
  \PackageWarning{mypackage}{prettyref is already loaded}
}{
  \usepackage{prettyref}
}
\@ifpackageloaded{cite}{
}{
  \usepackage{cite}
}
\usepackage{booktabs}
\usepackage{cellspace}
\makeatother
\usepackage{amsmath}
\usepackage{amsfonts}
\usepackage{amssymb}
\usepackage{multirow}
\usepackage{array}
\usepackage{textcomp}
\usepackage{stfloats}
\usepackage{verbatim}
\usepackage{graphicx}
\usepackage{mathtools}

\usepackage{longtable}

\usepackage{xcolor}

\usepackage{makecell}
\usepackage{algorithm}
\usepackage{algpseudocode}
\usepackage{tabularx}

\newrefformat{sec}{Sec.~\ref{#1}}
\newrefformat{ap:w
x}{Appendix~\ref{#1}}
\newrefformat{lem}{Lemma~\ref{#1}}
\newrefformat{fig}{Fig.~\ref{#1}}
\newrefformat{alg}{Algorithm~\ref{#1}}
\newrefformat{table}{Table~\ref{#1}}

\begin{document}

\title{ARMS: Anchor--Relational Motion Streaming for Seamless Solo-Social Motion Transitions}
\titlerunning{ARMS: Anchor--Relational Motion Streaming}

\author{Huakun Liu\inst{1}\orcidlink{0000-0002-9130-2519} \and
Qing Yu\inst{2}\orcidlink{0000-0001-6965-9581} \and
Kent Fujiwara\inst{2}\orcidlink{0000-0002-2205-6115} \and
Hideaki Uchiyama\inst{1}\orcidlink{0000-0002-6119-1184} \and
Kiyoshi Kiyokawa\inst{1}\orcidlink{0000-0003-2260-1707}}

\authorrunning{H.~Liu et al.}

\institute{Nara Institute of Science and Technology, Ikoma, Japan \and
LY Corporation, Tokyo, Japan \\
\email{\{liu.huakun.li0,hideaki.uchiyama,kiyo\}@is.naist.jp, \{yu.qing,kent.fujiwara\}@lycorp.co.jp}}

\maketitle

\begin{abstract}
Generating temporally continuous and socially coherent human motion from text remains a fundamental challenge, particularly in realistic streams where people act alone, enter interactions, and later disengage.
Most existing methods generate fixed-length motion clips under static agent configurations, which makes them brittle to solo--social transitions and unsuitable for incremental generation over long horizons.
We propose ARMS, an Anchor--Relational Motion Streaming framework that unifies solo motion and human--human interaction within a single causal generative process. 
ARMS introduces a dynamics-asymmetric representation that decouples per-person temporal evolution from inter-person alignment via a partner-referenced relative-translation term, enabling seamless switching of social coupling without sacrificing long-horizon stability or spatial consistency between agents.
On top of a causal latent space, a causal relational diffusion model progressively refines motion segment by segment using only past context, capturing both intra-person temporal dependencies and inter-person relations.
Mode-aware relational gating activates or masks cross-agent connections, allowing the same model to support both solo and interaction generation.
Experiments show that ARMS improves transition smoothness and social coherence compared to interaction-centric baselines, while also achieving competitive results on human--human interaction benchmarks.
Code is available at: \url{https://github.com/kk9six/arms}.
  
\end{abstract}

\section{Introduction}
\label{sec:intro}
\begin{figure}[tb]
  \centering
  \includegraphics[width=.97\textwidth]{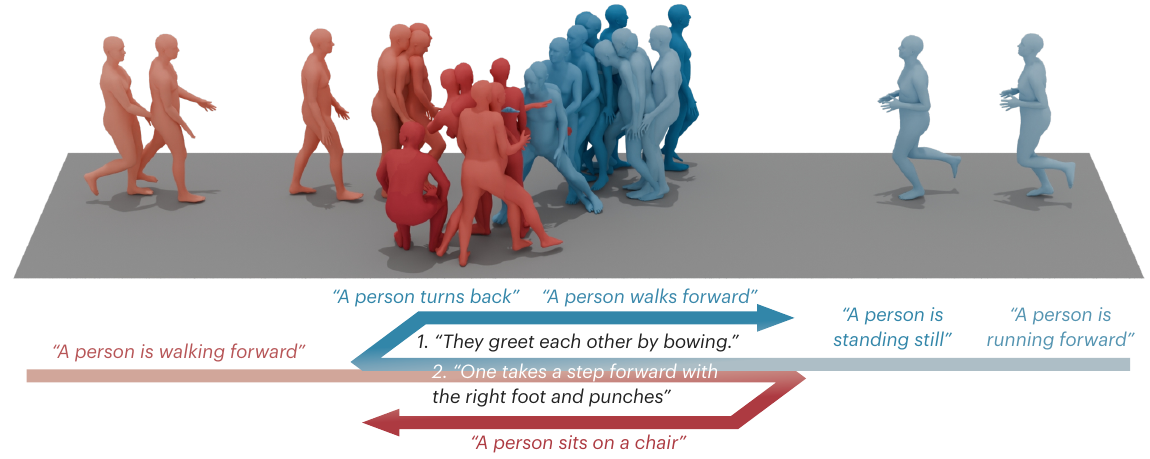}
  \caption{
  ARMS generates long-horizon motion streams from evolving textual instructions, enabling seamless transitions between solo behavior and human--human interaction within a single generative process. In this example, two individuals approach each other, engage in multiple interactions (e.g., greeting and fighting), and then disengage to continue with independent actions, without resetting or reinitializing generation.
  }
  \label{fig:overview}
\end{figure}
Text-driven human motion generation aims to synthesize human motion trajectories from natural language descriptions.
Most existing models treat motions as short and self-contained clips rather than continuous and socially coherent behaviors~\cite{sui2026survey,sahili2025text}.
However, in realistic settings, individuals may act alone, engage with others, or transition between solo and interactive behaviors~\cite{chensitcom, cai2024digital, lim2025event}.
The ability to generate such dynamic long-horizon behaviors is essential for applications including animation, virtual agents, robotics, and immersive environments.

Despite recent progress in human--human interaction generation, existing methods are typically designed to produce fixed-length motion clips in a single pass~\cite{sui2026survey}.
This design conflicts with realistic social behavior, where interactions emerge, evolve, and dissolve without fixed temporal boundaries.
Moreover, most approaches are tailored to a fixed agent configuration, modeling either solo motion or interaction using separate frameworks~\cite{cai2024digital,chensitcom,lim2025event}.
As a result, they struggle to support solo--social transitions in a temporally incremental setting, often causing boundary discontinuities or relational drift over long horizons.

Generating socially coherent motion in an incremental setting is challenging because the model must preserve temporal coherence for each individual while maintaining consistent inter-person relationships over extended horizons~\cite{liang2024intergen,cenready}.
Supporting streams that evolve between solo and two-person interaction further requires the model to accommodate changes in active agents and social coupling within a unified generative process.

To address these challenges, we propose \textbf{ARMS} (\textbf{A}nchor--\textbf{R}elational \textbf{M}otion \textbf{S}treaming), a causal autoregressive diffusion framework for temporally incremental human motion and interaction synthesis under changing one-/two-person social configurations.
ARMS is built on a novel \emph{dynamics-asymmetric representation} that unifies solo and interactive motions. The key is a partner-referenced relative-translation term that controls when inter-person coupling is represented. The term is inactive for solo motion and for the Anchor stream, but instantiated as the partner-relative displacement for the Relational stream during interaction. This design enables unbounded causal extension while avoiding global-coordinate drift, and preserves stable inter-person geometry when interactions emerge.

To enable streaming generation, motion sequences are first encoded into a temporally compressed latent space.
Crucially, the same causal temporal autoencoder encodes all individual motions, both in solo scenes and as components of interactions, under the proposed Anchor--Relational representation.
Building on a causal autoregressive diffusion backbone, we introduce \emph{mode-aware relational gating masks} that realize appropriate causal relational attention in both solo and interaction modes.
Dynamic history conditioning further improves long-horizon consistency by attending to variable-length past context.
Our contributions are:
\begin{itemize}
\item We formulate temporally incremental motion synthesis under \emph{one-/two-person social configuration changes}, unifying solo behavior and human--human interaction within a single streaming process that supports seamless solo--social transitions.
\item We propose a \emph{dynamics-asymmetric} Anchor--Relational motion representation with a partner-referenced relative-translation term, enabling configuration-agnostic learning from both solo and interaction data while using a shared causal temporal autoencoder for individual motion encoding.
\item Building on an autoregressive diffusion backbone, we instantiate the denoiser as a causal relational transformer and introduce \emph{mode-aware relational gating masks} with dynamic history conditioning, enabling incremental generation while preserving temporal coherence and stable inter-person geometry.
\end{itemize}

\section{Related Work}
\label{sec:related-work}
Human motion generation has been widely studied under various conditioning modalities, including action labels~\cite{guo2020action2motion,petrovich2021action}, audio signals~\cite{alexanderson2023listen,liu2024emage,ghosh2025duetgen, chen2021choreomaster, li2021ai}, scene context~\cite{mullen2023placing, zhao2023synthesizing, lim2025event, chensitcom,tevetclosd}, and natural language~\cite{dabral2023mofusion, guo2022generating, guo2022tm2t, jiang2023motiongpt, zhu2025motiongpt3,petrovich2022temos,tevet2022human}.
We focus on text conditioning for its flexibility and accessibility~\cite{zhu2025motiongpt3, ouyang2025motion, chensitcom}.

\noindent\textbf{Human--Human Interaction Generation.}
\label{sec:rw-interaction}
Early motion generation primarily focused on single-person dynamics without social interactions.
Building upon a pretrained single-person diffusion model, ComMDM~\cite{shafirhuman} extends it to two-person generation by learning a lightweight communication module, while RIG~\cite{tanaka2023role} and InterGen~\cite{liang2024intergen} introduce cooperative diffusion denoisers with two shared-weight Transformer branches.
To enhance interaction diversity, in2IN~\cite{ruiz2024in2in} and MixerMDM~\cite{ruiz2025mixermdm} augment interaction modeling with additional single-person motion data, while Text2Interact~\cite{wu2025text2interact} introduces a scalable synthesis-by-composition pipeline with fine-grained word-level conditioning.
TIMotion~\cite{wang2025timotion} explicitly structures interaction dynamics along a temporal sequence to better capture evolving inter-person dependencies.
Beyond continuous diffusion, InterMask~\cite{javedintermask} formulates interactions in a discrete 2D VQ space with masked modeling, and SocialGen~\cite{yusocialgen} aligns motion and language tokens to extend interaction modeling toward multi-human social scenarios.

Despite these advances, many interaction models remain optimized for fixed-length synthesis from a static prompt, and do not support incremental extension. Moreover, most interaction methods assume a fixed two-person setting and do not address solo--social mode switching in evolving streams, where agents may approach, interact, and disengage without restarting generation.
This leaves open the problem of unified streaming motion synthesis that preserves both per-person temporal coherence and stable inter-person geometry across transitions.

\noindent\textbf{Long-horizon Motion Generation.}
\label{sec:rw-streaming-varcard}
Recent works extend motion generation beyond isolated short clips toward long-horizon or sequential synthesis~\cite{petrovich2024multi, barquero2024seamless, zhuo2025infinidreamer, xiao2025motionstreamer, zhaodartcontrol, zhang2025primal, tevetclosd}.
Several methods perform autoregressive diffusion for real-time control, generating motion incrementally from past states, including CAMDM~\cite{chen2024taming}, A-MDM~\cite{shi2024interactive}, and PRIMAL~\cite{zhang2025primal}.
CLoSD~\cite{tevetclosd} and DART~\cite{zhaodartcontrol} formulate motion as autoregressive segments, while InfiniDreamer~\cite{zhuo2025infinidreamer} explores arbitrarily long motion synthesis via segment refinement and score distillation.
MotionStreamer~\cite{xiao2025motionstreamer} operates in a continuous causal latent space with a temporally causal autoencoder and strictly causal masking.
Most long-horizon autoregressive methods are developed for single-person motion generation~\cite{guo2022generating}.
Recent interaction extensions such as Interact2Ar~\cite{ruiz2025interact2ar} and HINT~\cite{liu2026hint} extend diffusion-based interaction models to the autoregressive setting via sliding-window generation. Interact2Ar incorporates a memory mechanism to reuse past motion features, while HINT adopts hierarchical conditioning in a canonicalized latent space, which can introduce discontinuities across segments.

Overall, existing approaches either generate human--human interactions within fixed-length clips, or extend motion synthesis to long horizons primarily for single-person scenarios.
Autoregressive interaction extensions typically still assume a fixed agent configuration and focus on maintaining continuity within that regime.
Consequently, streaming generation under changing one-/two-person social configurations remains largely unexplored.

\noindent\textbf{Interactive Motion Representation.}
Motion representation is critical for both motion quality and long-horizon stability in human motion generation.
Most single-person motion generation methods adopt canonicalized representations derived from HumanML3D~\cite{guo2022generating}, expressing motion in a root-centered coordinate system with frame-to-frame relative updates.
Such incremental formulations simplify learning and naturally support temporally extensible generation~\cite{tevetclosd,xiao2025motionstreamer}.
Several interaction models extend this paradigm by introducing initial relative transformations between agents~\cite{shafirhuman, yusocialgen}, but accumulated incremental updates can introduce spatial drift that disrupts inter-person alignment~\cite{liang2024intergen}.
To avoid this issue, InterGen~\cite{liang2024intergen} models motion in global coordinates to explicitly preserve spatial relations between agents, which improves alignment but constrains motion generation to the coordinate distributions observed during training and limits their compatibility with open-ended or temporally extensible synthesis.
HINT~\cite{liu2026hint} mitigates drift by generating short motion segments in a canonicalized latent space, but segment-wise canonicalization introduces discontinuities that cause jittery transitions.
DuetGen~\cite{ghosh2025duetgen} instead represents two-person motion as a unified long feature sequence that jointly encodes canonicalized motion and relational features, but is limited to fixed two-person scenarios and increases dimensionality.

These designs highlight a fundamental tension in interaction representation: maintaining stable inter-person spatial relations while enabling temporally extensible generation. 
A shared representation that supports causal encoding and autoregressive rollout, preserves interaction geometry, and enables seamless solo--social transitions within a unified streaming process remains underexplored.

\section{Method}
\label{sec:method}

\subsection{Problem Formulation}
\label{sec:problem}
As shown in~\cref{fig:overview}, we address the task of streaming motion generation under changing one-/two-person social configurations, where textual instructions arrive incrementally, with one text remaining active until replaced by a new one.
Let $\mathcal{T} = \{\tau_1, \tau_2, \dots\}$ denote the ordered instruction stream, and $\beta(t)$ the index of the instruction active at timestep $t$, inducing change points that partition the timeline into variable-length intervals.
At each timestep $t$, the model generates the motion states of all active agents:
\begin{equation}
  \{\mathbf{x}_t^{(i)}\}_{i=1}^{N_t},
\end{equation}
where $N_t \in \{1, 2\}$ denotes the number of active agents at time $t$.
We denote the generated motion history before $t$ as
\begin{equation}
\mathcal{H}_{<t} = 
\{
\mathbf{x}_{t'}^{(i)} \mid t' < t,;i = 1,\dots,N_{t'}
\},
\end{equation}
where the number of active agents in the history may vary over time.  
The objective is therefore to learn a causal generative model that produces motion incrementally,
\begin{equation}
  p\big( \{\mathbf{x}_t^{(i)}\}_{i=1}^{N_t}
    \mid
    \mathcal{H}_{<t},
    \tau_{\beta(t)},
    N_t
  \big),
\end{equation}
while preserving long-horizon temporal coherence within each individual and maintaining relational consistency whenever interactions are present.

\begin{figure}[tb]
  \centering
  \includegraphics[width=.95\textwidth]{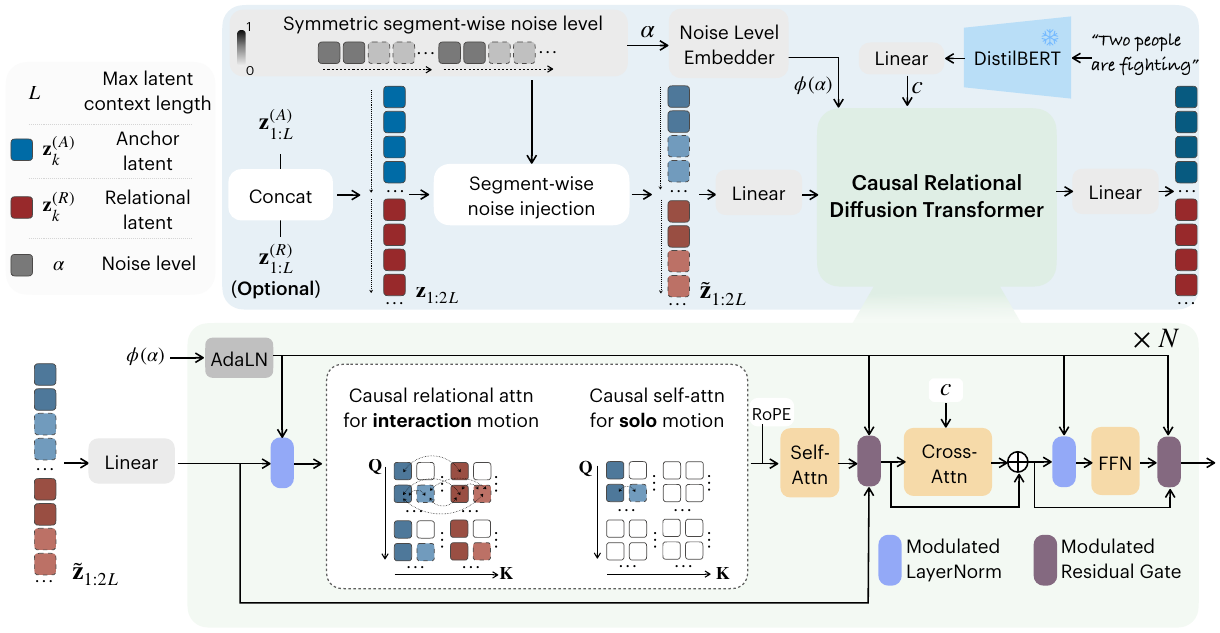}
  \caption{We encode motion into causal latent streams for the Anchor and Relational agents, concatenate them, and apply symmetric segment-wise noising.
  A text-conditioned Causal Relational Diffusion Transformer denoises the sequence using segment-wise causal attention and mode-aware relational gating masks, enabling cross-agent visibility for interaction and masking it for solo motion.
  }
  \label{fig:fig_2_overview}
\end{figure}

\subsection{Motion Representation with Asymmetric Dynamics}
\label{sec:representation}
Streaming motion generation commonly adopts canonicalized incremental features that enable temporally extensible synthesis, but these representations can accumulate spatial drift over long horizons.  
In contrast, interaction models often operate in global coordinates to preserve inter-person alignment, at the cost of restricting generation to coordinate distributions observed during training.
To balance these trade-offs, inspired by~\cite{liang2024intergen, ghosh2025duetgen}, we adopt a dynamics-asymmetric representation that separates incremental temporal dynamics from relational dynamics within a shared motion state.
The key is a partner-referenced relative-translation term that controls when inter-person coupling is represented: it is set to zero for solo motion (and for the Anchor stream), and instantiated as the relative displacement to the partner for the Relational stream during interaction.
This design preserves temporally extensible causal rollout while maintaining stable inter-person spatial geometry whenever interactions are present.

\noindent\textbf{Per-frame state definition.}
We adopt a coordinate system with axes defined as $x$-left, $y$-up, and $z$-forward.
At each timestep $t$, the motion state of an agent is defined as
\begin{equation}
  \mathbf{x}_t =
  \left[
    \mathbf{r}_t^{yaw},
    \mathbf{v}_t^{xz},
    \mathbf{\Delta}_t^{xz},
    \mathbf{p}_t^{local},
    \dot{\mathbf{p}}_t^{local},
    \mathbf{q}_t,
    \mathbf{c}_t
  \right],
\end{equation}
where $\mathbf{r}_t^{yaw} = [\sin(\theta_t), \cos(\theta_t)] \in \mathbb{R}^2$ encodes the root yaw orientation to stabilize long-horizon integration and avoid angular discontinuities.
The root linear velocity $\mathbf{v}_t^{xz} \in \mathbb{R}^2$ and the relational displacement $\mathbf{\Delta}_t^{xz} \in \mathbb{R}^2$ are defined on the world-frame $xz$-plane.
Joint positions $\mathbf{p}_t^{local} \in \mathbb{R}^{J\times 3}$ and velocities $\dot{\mathbf{p}}_t^{local} \in \mathbb{R}^{J\times 3}$ are expressed in the root-local frame.
Joint rotations $\mathbf{q}_t \in \mathbb{R}^{J\times 6}$ adopt the continuous 6D rotation representation~\cite{zhou2019continuity} for all $J$ non-root joints.
Foot-contact indicators $\mathbf{c}_t \in \mathbb{R}^{4}$ encode binary contact states for four predefined foot joints.

\noindent\textbf{Asymmetric motion dynamics.}
Although both agents share the same state formulation, the two motion dynamics are instantiated asymmetrically during generation.
We designate one agent as the \textit{Anchor Agent (A)}, which models canonical temporal dynamics, and the other as the \textit{Relational Agent (R)}, which explicitly encodes relational alignment.
For the Anchor branch, the relational displacement is fixed:
\begin{equation}
  \mathbf{\Delta}_t^{(A)} = \mathbf{0},
\end{equation}
which ensures that the Anchor motion is modeled independently of inter-agent spatial relations and remains compatible with solo motion generation.
The global root position $\mathbf{P}_t^{(A)}$ is obtained via integration of $\mathbf{v}_t^{xz}$ over time.
The Relational branch encodes interaction dynamics.
It stores the root translation offset $\mathbf{\Delta}_t^{(R)}$ with respect to the Anchor agent.
Its global root position is reconstructed as
\begin{equation}
  \mathbf{P}_t^{(R)} = \mathbf{P}_t^{(A)} + \mathbf{\Delta}_t^{(R)}.
\end{equation}
Restricting global integration to a single branch prevents independent integration errors that cause drifting global trajectories and unstable inter-agent alignment.
Both agents retain identical state space, which enables architectural symmetry and parameter sharing. 
Notably, when the relational displacement is inactive, the Anchor branch alone reduces to canonical incremental motion modeling, making the formulation naturally compatible with solo motion generation.

\subsection{Causal Temporal AutoEncoder}
\label{sec:tae}
Long-horizon autoregressive generation in high-dimensional motion space is computationally costly and prone to drift. Following the causal compression paradigm introduced in~\cite{xiao2025motionstreamer}, we adopt a \textit{causal variational autoencoder} to compress motion into a temporally causal latent space, where each latent depends only on past frames and supports incremental decoding during streaming generation.
The model is a temporal variational autoencoder composed of 1D causal convolution and residual blocks in both encoder and decoder.
Given a motion sequence ${\mathbf{x}_t}_{t=1}^{T}$ for a single agent, the encoder produces temporally downsampled Gaussian parameters ${\boldsymbol{\mu}_k, \boldsymbol{\sigma}_k^2}_{k=1}^{T/l}$, where $l$ is the temporal downsampling factor.
Latent variables $\mathbf{z}_k \in \mathbb{R}^d$ are sampled via reparameterization, forming a continuous causal latent sequence $\mathbf{z}_{1:T/l}$.
For latent timestep $k$, we denote by $\mathbf{z}_{\le k}$ the causal latent prefix available up to that timestep.
The decoder reconstructs motion from the causal latent prefix $\mathbf{z}_{\le k}$.

For interaction modeling, the same autoencoder is applied independently to each agent branch with shared weights.
The motion of each individual is encoded identically whether it appears in a solo sequence or as part of an interaction under our Anchor--Relational representation.
This yields a unified latent space while preserving independent temporal encoding for different agents.

\subsection{Causal Relational Diffusion}
\label{sec:dit}
Let $\mathbf{z}_k^{(A)} \in \mathbb{R}^{d}$ and $\mathbf{z}_k^{(R)} \in \mathbb{R}^{d}$ denote the causal latents at latent timestep $k$ for the Anchor and Relational agent, respectively.
For a window of length $L$ latent steps per agent, where $L$ denotes the maximum context size, we form a single ordered latent stream by concatenating the two agent streams into a single sequence:
\begin{equation}
  \mathbf{z}_{1:2L} =
  \big[\mathbf{z}^{(A)}_{1:L}\,;\,\mathbf{z}^{(R)}_{1:L}\big].
\end{equation}

\noindent\textbf{Segment-wise causal noising in latent space.}
Following diffusion forcing~\cite{chen2024diffusion}, we adopt a temporally structured noise schedule such that earlier latent segments are assigned lower noise levels than later ones.
This creates a progressive certainty gradient along the temporal axis, encouraging committed refinement of past frames while preserving flexibility for future ones.
Instead of assigning independent noise levels for every latent timestep, we partition the latent timeline into contiguous non-overlapping segments of size $S$, where $S$ denotes the number of latent steps in each segment.
All latent steps within the same segment share a common noise level.
Let $s(k)=\lceil k/S\rceil$ denote the segment index for timestep $k$, and let $\alpha_{s}\in[0,1]$ denote the noise level associated with segment $s$.
The noised latent at timestep $k$ is constructed as
\begin{equation}
  \tilde{\mathbf{z}}_{k} = \alpha_{s(k)}\,\mathbf{z}_{k} + (1-\alpha_{s(k)})\,\boldsymbol{\epsilon}_{k},
  \qquad \boldsymbol{\epsilon}_{k}\sim\mathcal{N}(\mathbf{0},\mathbf{I}).
  \label{eq:segmentwise_noising}
\end{equation}
Segment-wise noise sharing enforces joint refinement over short temporal segments rather than isolated latent steps.
This increases the effective temporal context at each diffusion stage and improves local motion consistency.
For the two-agent latent stream $\mathbf{z}_{1:2L}$, segments corresponding to the same timestep of each agent share identical noise levels.
This symmetric assignment ensures that the Anchor and Relational branches are refined at the same temporal stage, which stabilizes interaction dynamics during diffusion.

\noindent\textbf{Relational diffusion denoiser.}
The denoiser is built upon the causal diffusion architecture proposed in~\cite{yu2026causal}.
As shown in~\cref{fig:fig_2_overview}, the noised latent stream $\tilde{\mathbf{z}}_{1:2L}$ is processed by a stack of transformer layers that jointly model temporal dynamics and inter-agent relations under a segment-wise causal attention mechanism.
Noise levels are first embedded and injected into each transformer layer through timestep-conditioned adaptive layer normalization (AdaLN)~\cite{peebles2023scalable}.
Self-attention is then applied with rotary positional encoding (RoPE)~\cite{su2024roformer} along the temporal dimension to preserve ordering under incremental extension.
To enforce temporal causality and relational reasoning, we apply a segment-wise causal relational attention mask to the self-attention logits.

Let $q,r \in \{1,\dots,2L\}$ index query and key positions in the concatenated latent stream.
For any position $u$, we denote by $a(u)\in\{A,R\}$ its agent identity and by
$k(u)\in\{1,\dots,L\}$ its per-agent latent timestep.
Timesteps are grouped into segments of size $S$ with segment index
\begin{equation}
  s(u)=\left\lceil \frac{k(u)}{S}\right\rceil.
\end{equation}
The relational attention mask is defined as
\begin{equation}
  M_{qr} =
  \begin{cases}
    0, & s(r)\le s(q)\ \wedge\ \Gamma\!\big(a(q),a(r)\big)=1, \\
    -\infty, & \text{otherwise}\,.
  \end{cases}
\end{equation}
This formulation ensures that each query attends only to its own, past segments, and optionally cross-agent segments, preventing future leakage. $\Gamma(\cdot,\cdot)$ is a mode gate that controls cross-agent visibility and is defined as
\begin{equation}
  \Gamma(a(q),a(r))=
  \begin{cases}
    1, & \text{interaction mode}, \\
    1[a(q)=a(r)=A], & \text{solo mode}.
  \end{cases}
\end{equation}
In solo mode, only the Anchor tokens remain visible, while Relational tokens are ignored.
After the self-attention, text instructions are encoded using a frozen DistilBERT encoder~\cite{sanh2019distilbert}, producing token embeddings that are linearly projected to the model dimension and incorporated via cross-attention in every transformer layer.
This design allows latent vectors to condition on the active textual instruction while maintaining temporal causality in the motion stream.
The overall denoiser predicts velocity-like residuals~\cite{meng2025absolute} for all latent steps jointly:
\begin{equation}
  \mathbf{v}_\theta
  =
  \mathcal{F}_\theta
  \!\left(
    \tilde{\mathbf{z}}_{1:2L},
    \phi(s),
    \mathbf{c},
    M
  \right),
\end{equation}
where $\phi(s)$ denotes embedded segment noise levels, $\mathbf{c}$ the text features, and $M$ the mode-aware relational gating mask.

\noindent\textbf{Inference.}
At inference time, we generate motions by progressively refining latent segments under a temporally offset noise schedule~\cite{chen2024diffusion, yu2026causal}.
Let $K$ denote the number of diffusion refinement iterations and $\delta$ the refinement offset between adjacent segments.
Segment $i+1$ begins denoising $\delta$ refinement steps after segment $i$.
Under this schedule, earlier segments become deterministic sooner while later segments remain partially noisy and continue to be refined.
This overlapping denoising avoids hard segment boundaries and improves efficiency compared to strictly sequential autoregressive refinement.
For continuous generation, we adopt variable-length history conditioning during inference.
Under the same prompt, the full context window $L$ is preserved to maintain long-range temporal coherence.
When the textual instruction changes, we retain only a short history $H$ of previously generated latents while replacing the conditioning text embedding $\mathbf{c}$ for future segments.
This reduced history provides sufficient motion context for transition while allowing the model to adapt smoothly to the new instruction.
Agent configurations can also change dynamically during inference.
When generating solo motion, the Relational branch is masked through $\Gamma(\cdot,\cdot)$, reducing the model to a single-agent generator.
When transitioning to interaction, the Relational branch is activated and initialized either from noise or from encoded relative dynamics derived from existing motion. Unless otherwise stated, we initialize the Relational branch from noise at interaction onset.
Details of the exact inference procedure are provided in the supplementary material.

\section{Experiments}
\label{sec:experiments}


\subsection{Experimental Setup}
\noindent\textbf{Dataset.}
\label{sec:exp-dataset}
We use HumanML3D~\cite{guo2022generating} for single-person motion and InterHuman~\cite{liang2024intergen} for two-person interactions.
We adopt a 272-dimensional representation variant of the HumanML3D dataset used in~\cite{xiao2025motionstreamer} for better compatibility with InterHuman.
It contains 26{,}846 motion sequences derived from the AMASS~\cite{mahmood2019amass} dataset, each paired with three textual descriptions.
InterHuman consists of 7{,}779 two-person interaction sequences with 23{,}337 textual descriptions.
Since HumanML3D and InterHuman adopt different motion representations, we convert both datasets into our shared Anchor--Relational representation for training.
This conversion is deterministic and invertible.
For evaluation, generated motions are converted back to the native representation of each dataset before computing metrics, strictly following standard evaluation protocols.
We train a unified model jointly on HumanML3D and InterHuman for cross-scenario streaming motion generation, while training dataset-specific models on InterHuman for fair comparison in interaction generation evaluation.

Following prior work~\cite{javedintermask}, we additionally report results on InterX~\cite{xu2024inter} to assess cross-skeleton generalization.
InterX follows the SMPL-X~\cite{pavlakos2019expressive} representation and contains 11{,}388 motion sequences, each paired with three textual descriptions. To ensure compatibility with our model, we convert the InterX motions into the same anchor--relational representation used for training.

\noindent\textbf{Implementation details.}
\label{sec:exp-imple}
The causal temporal encoder downsamples the motion sequences by a factor of $4$, producing latent tokens with dimension $64$.
The motion generation model is implemented as a causal relational diffusion transformer with $8$ layers, a hidden dimension of $512$, $4$ attention heads, a maximum context length of $L=75$, and a segment size of $S = 5$.
Text conditions are extracted using a frozen DistilBERT encoder.
We train the denoiser with a batch size of $64$ and a maximum sequence length of $300$ frames, i.e., 75 latent tokens, for $500$ epochs.
Optimization uses AdamW with learning rate $2\times 10^{-4}$, betas $(0.9,0.99)$, and weight decay $1\times 10^{-5}$.
We employ flow matching~\cite{albergobuilding,albergo2025stochastic,meng2025absolute} as the ODE sampler in causal diffusion forcing.
During inference, we perform $K=50$ denoising steps, and the refinement offset $\delta$ is set to $5$.

\subsection{Quantitative Evaluation on Interaction Generation}
\begin{table}[tb]
  \caption{\textbf{Quantitative evaluation} on the \textbf{InterHuman} test sets.
  Values are reported with 95\% confidence intervals.
  The arrow $\rightarrow$ indicates that performance closer to ground truth is preferred.
  Best and second-best are \textbf{bold} and \underline{underlined}.}
  \label{table:interhuman-eval}
  \centering
  \resizebox{\linewidth}{!}{
  \begin{tabular}{l c c c c c c c }
    \toprule
    \multirow{2.5}{*}{Method} &
    \multicolumn{3}{c}{R Precision\(\uparrow\)} &
    \multirow{2.5}{*}{FID\(\downarrow\)} &
    \multirow{2.5}{*}{MM Dist\(\downarrow\)} &
    \multirow{2.5}{*}{Diversity\(\rightarrow\)} \\
    \cmidrule{2-4}
    & Top 1 & Top 2 & Top 3 & & & & \\
    \midrule
    Ground Truth &
    0.452$^{\pm.008}$ &
    0.610$^{\pm.009}$ &
    0.701$^{\pm.008}$ &
    0.273$^{\pm.007}$ &
    3.755$^{\pm.008}$ &
    7.948$^{\pm.064}$ \\
    \midrule
    T2M~\cite{guo2022generating} &
    0.238$^{\pm.012}$ &
    0.325$^{\pm.010}$ & 
    0.464$^{\pm.014}$ &
    13.769$^{\pm.072}$ &
    5.731$^{\pm.013}$ &
    7.046$^{\pm.022}$ \\

    MDM~\cite{tevethuman}&
    0.153$^{\pm.012}$&
    0.339$^{\pm.012}$&
    0.339$^{\pm.012}$ &
    9.167$^{\pm.056}$ &
    7.125$^{\pm.018}$ &
    7.602$^{\pm.045}$ \\

    ComMDM~\cite{shafirhuman}&
    0.223$^{\pm.009}$&
    0.334$^{\pm.008}$&
    0.466$^{\pm.010}$&
    7.069$^{\pm.054}$&
    6.212$^{\pm.021}$ &
    7.244$^{\pm.038}$ \\

    RIG~\cite{tanaka2023role}&
    0.285$^{\pm.010}$&
    0.409$^{\pm.014}$&
    0.521$^{\pm.013}$ &
    6.775$^{\pm.069}$ &
    4.876$^{\pm.018}$ &
    7.311$^{\pm.043}$ \\

    InterGen~\cite{liang2024intergen} &
    0.371$^{\pm.010}$&
    0.515$^{\pm.012}$&
    0.624$^{\pm.010}$ &
    5.918$^{\pm.079}$ &
    5.108$^{\pm.014}$ &
    7.387$^{\pm.029}$ \\

    MoMat–MoGen~\cite{cai2024digital} &
    0.449$^{\pm.004}$&
    0.591$^{\pm.003}$&
    0.666$^{\pm.004}$ &
    5.674$^{\pm.085}$ &
    3.790$^{\pm.001}$ &
    8.021$^{\pm.350}$ \\

    in2IN~\cite{ruiz2024in2in} &
    0.425$^{\pm.008}$ &
    0.576$^{\pm.008}$ &
    0.662$^{\pm.009}$ &
    5.535$^{\pm.120}$ & 3.803$^{\pm.002}$ & 7.953$^{\pm.047}$\\
    InterMask~\cite{javedintermask} &
    0.449$^{\pm.004}$&
    0.599$^{\pm.005}$&
    0.683$^{\pm.004}$ &
    5.154$^{\pm.061}$ &
    3.790$^{\pm.002}$ &
    {\bf 7.944$^{\pm.033}$} \\
    Text2Interact~\cite{wu2025text2interact} &
    0.483$^{\pm.005}$&
    0.638$^{\pm.005}$&
    0.717$^{\pm.005}$ &
    5.191$^{\pm.055}$ &
    3.778$^{\pm.001}$ &
    7.900$^{\pm.030}$ \\
    HINT~\cite{liu2026hint} &
    0.432$^{\pm.004}$&
    0.587$^{\pm.004}$&
    0.672$^{\pm.004}$ &
    {\bf 3.100}$^{\pm.035}$ &
    3.796$^{\pm.001}$ &
    7.898$^{\pm.023}$ \\
    TIMotion~\cite{wang2025timotion} &
    \underline{0.501}$^{\pm.005}$&
    \underline{0.656}$^{\pm.006}$&
    \underline{0.734}$^{\pm.006}$ &
    4.702$^{\pm.069}$ &
    \underline{3.769}$^{\pm.001}$ &
    \underline{7.943}$^{\pm.034}$ \\
    \midrule
    Ours (full-window generation) &
    {\bf0.529}$^{\pm.008}$&
    {\bf0.690}$^{\pm.008}$&
    {\bf 0.764$^{\pm.004}$} &
    {\underline 4.436$^{\pm.069}$} &
    {\bf 3.763$^{\pm.002}$} &
    8.089$^{\pm.059}$ \\
    Ours (streaming generation)  &
    0.492$^{\pm.004}$&
    0.647$^{\pm.005}$&
    0.723$^{\pm.005}$ &
    4.444$^{\pm.068}$ &
    3.778$^{\pm.001}$ &
    8.017$^{\pm.026}$ \\
    \bottomrule
  \end{tabular}
  }
\end{table}

We evaluate human--human interaction generation capability of ARMS under the standard benchmark setting.
We follow evaluation protocols in text-to-motion generation using the pretrained text--motion retrieval evaluator adopted in prior work~\cite{liang2024intergen,javedintermask}.
We report: (1) R-Precision, the retrieval accuracy of the ground-truth motion given a text query; (2) MM Dist, the average embedding distance between generated motions and their paired captions; (3) FID, the Fr\'echet distance between feature distributions of generated and real motions; and (4) Diversity, the average pairwise distance among generated motions.
All methods are evaluated under identical settings as prior work for fair comparison.

We evaluate our model under two inference regimes.
\emph{Full-window generation} performs diffusion over the entire motion window in a single denoising process.
\emph{Streaming generation} adopts our causal segment-wise refinement schedule and incrementally synthesizes motions.
Table~\ref{table:interhuman-eval} reports quantitative results on the InterHuman test set. 
Our \emph{full-window generation} variant achieves the best overall performance among all compared methods, indicating superior text--motion alignment and realism.
In the \emph{streaming generation} setting, ARMS maintains strong quality (FID $4.44$, MM Dist $3.78$), with lower R-Precision due to compounding uncertainty over long incremental rollouts.
Importantly, streaming generation enables long-duration motion synthesis beyond fixed-length clips while maintaining comparable realism and alignment quality.

\begin{table}[tb]
  \caption{\textbf{Quantitative evaluation} on the \textbf{InterX} test sets.
  Values are reported with 95\% confidence intervals.
  The arrow $\rightarrow$ indicates that performance closer to ground truth is preferred.
  Best and second-best results are {\bf bold} and \underline{underlined}.}
  \label{table:interx-eval}
  \centering
  \resizebox{\linewidth}{!}{
  \begin{tabular}{l c c c c c c c }
    \toprule
    \multirow{2.5}{*}{Method} &
    \multicolumn{3}{c}{R Precision\(\uparrow\)} &
    \multirow{2.5}{*}{FID\(\downarrow\)} &
    \multirow{2.5}{*}{MM Dist\(\downarrow\)} &
    \multirow{2.5}{*}{Diversity\(\rightarrow\)} \\
    \cmidrule{2-4}
    & Top 1 & Top 2 & Top 3 & & & & \\
    \midrule
    Ground Truth &
    0.429$^{\pm.004}$&
    0.626$^{\pm.003}$&
    0.736$^{\pm.003}$ &
    0.002$^{\pm.0002}$ &
    3.536$^{\pm.013}$ &
    9.734$^{\pm.078}$ \\
    \midrule
    T2M~\cite{guo2022generating} & 
    0.184$^{\pm.010}$&
    0.298$^{\pm.006}$&
    0.396$^{\pm.005}$ &
    5.481$^{\pm.382}$ &
    9.576$^{\pm.006}$ &
    2.771$^{\pm.151}$ \\

    MDM~\cite{tevethuman} &
    0.203$^{\pm.009}$&
    0.329$^{\pm.007}$&
    0.426$^{\pm.005}$ &
    23.701$^{\pm.057}$ &
    9.548$^{\pm.014}$ &
    5.856$^{\pm.077}$ \\

    ComMDM~\cite{shafirhuman} &
    0.090$^{\pm.002}$&
    0.165$^{\pm.004}$&
    0.236$^{\pm.004}$ &
    29.266$^{\pm.067}$ &
    6.870$^{\pm.017}$ &
    4.734$^{\pm.067}$ \\

    InterGen~\cite{liang2024intergen} &
    0.207$^{\pm.004}$&
    0.335$^{\pm.005}$&
    0.429$^{\pm.005}$ &
    5.207$^{\pm.216}$ & 9.580$^{\pm.011}$ & 7.788$^{\pm.208}$ \\

    InterMask~\cite{javedintermask} &
    0.403$^{\pm.005}$&
    0.595$^{\pm.004}$&
    0.705$^{\pm.005}$&
    0.399$^{\pm.013}$&
    3.705$^{\pm.017}$&
    9.046$^{\pm.073}$ \\
    TIMotion~\cite{wang2025timotion} &
    0.412$^{\pm.004}$&
    0.601$^{\pm.004}$&
    0.714$^{\pm.003}$ &
    0.385$^{\pm.022}$ &
    3.706$^{\pm.015}$ &
    \underline{9.191}$^{\pm.092}$ \\
    Interact2Ar~\cite{ruiz2025interact2ar} &
    \underline{0.441}$^{\pm.00}$&
    \underline{0.631}$^{\pm.00}$&
    \underline{0.737}$^{\pm.00}$&
    {\bf0.148}$^{\pm.01}$&
    \underline{3.581}$^{\pm.01}$&
    9.147$^{\pm.06}$\\
    HINT~\cite{liu2026hint} &
    0.386$^{\pm.005}$&
    0.572$^{\pm.004}$&
    0.682$^{\pm.003}$&
    \underline{0.278}$^{\pm.012}$&
    4.007$^{\pm.016}$&
    8.886$^{\pm.066}$\\
    \midrule
    Ours (streaming generation) &
    {\bf0.479}$^{\pm.005}$ &
    {\bf0.679}$^{\pm.005}$&
    {\bf0.780$^{\pm.003}$} &
    0.279$^{\pm.010}$ &
    {\bf3.405$^{\pm.016}$} &
    {\bf9.399$^{\pm.064}$} \\
    \bottomrule
  \end{tabular}
  }
\end{table}
Table~\ref{table:interx-eval} presents results on InterX.
Our streaming generation model outperforms other models on all metrics.
Notably, the improved R precision and diversity demonstrates the strong text--motion alignment and rich motion variation.
These results indicate that our formulation generalizes effectively across different skeletal definitions, without being specialized to a specific joint configuration.

%
\subsection{Motion Streaming: Qualitative Evaluation}
\begin{figure}[tb]
  \centering
  \includegraphics[width=.95\textwidth]{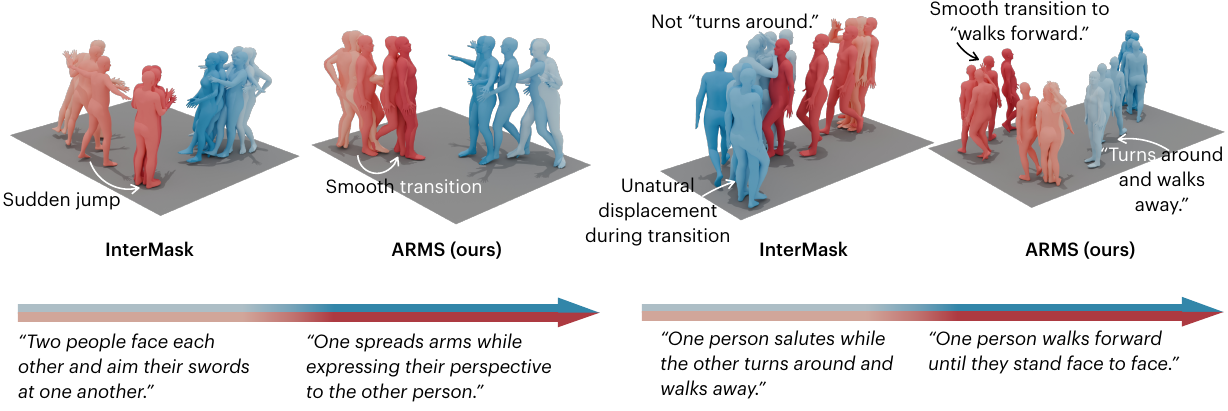}
  \caption{Qualitative comparison of long-duration interaction synthesis.
  Compared to InterMask, our method generates smoother temporal transitions, avoids abrupt motion discontinuities, and preserves coherent interaction dynamics across extended sequences.}
  \label{fig:interaction_comparison}
\end{figure}
\begin{figure}[tb]
  \centering
  \includegraphics[width=.95\textwidth]{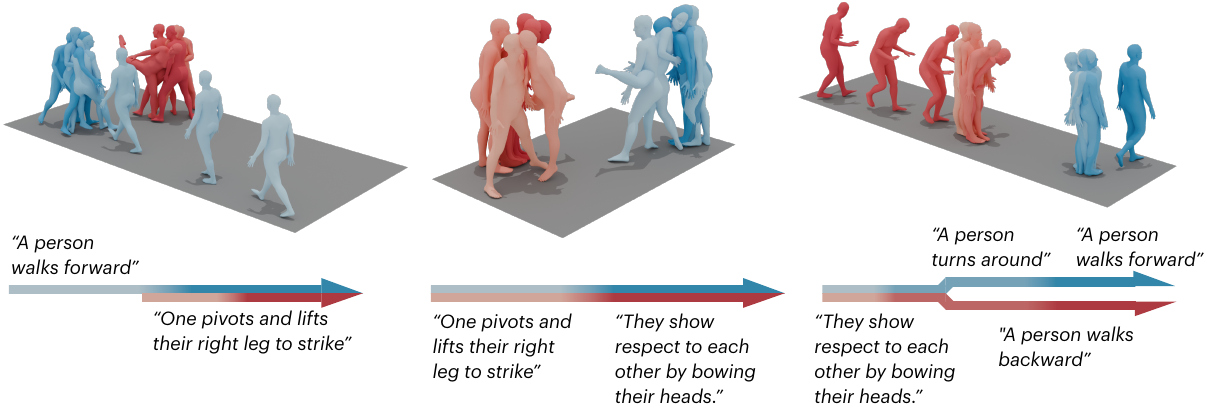}
  \caption{Qualitative results of streaming motion generation.
  From left to right, we show: (1) solo motion transitioning into human--human interaction, (2) interaction evolving into a different interaction, and (3) interaction resolving back into solo motion.
  All sequences are generated in a streaming manner without resetting the latent state.}
  \label{fig:multiple_scenarios}
\end{figure}
\begin{figure}[tb]
  \centering
  \includegraphics[width=.95\textwidth]{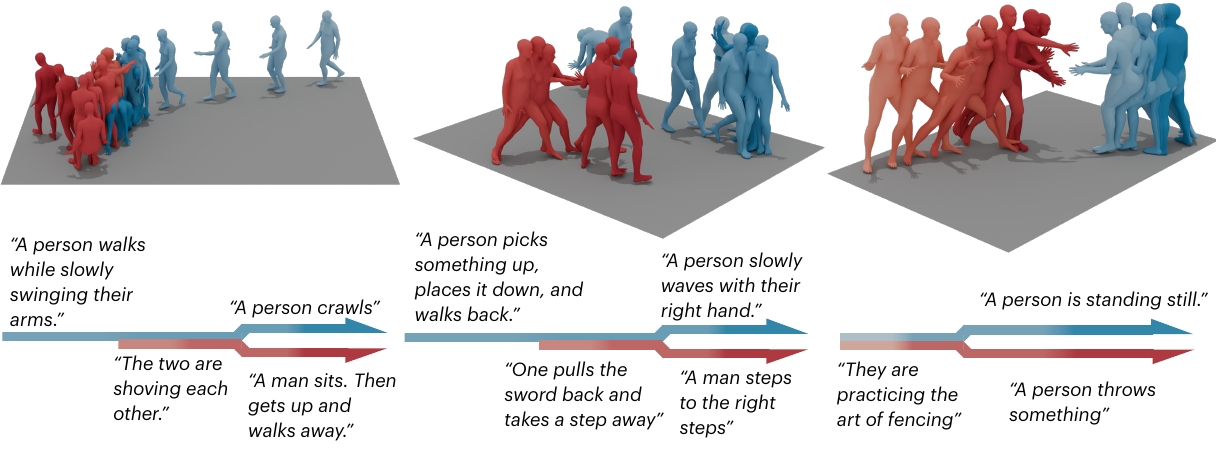}
  \caption{Qualitative results of dynamic social motion streaming generation.}
  \label{fig:more-results-transition}
\end{figure}
We present qualitative results on long-duration interaction synthesis and cross-scenario generalization.
\cref{fig:interaction_comparison} compares our method with the adapted InterMask.
In extended sequences, InterMask often exhibits unnatural spatial shifts during interaction transition.
In contrast, our method achieves smooth transitions between sub-actions, maintains consistent relative positioning, and preserves semantically aligned interaction evolution over time, demonstrating improved stability in long-duration interaction synthesis.
\cref{fig:multiple_scenarios}~and~\cref{fig:more-results-transition} further illustrate results of three cross-agent scenarios: (1) a solo motion transitions into a human--human interaction; (2) an interaction evolves into another; and (3) an interaction resolves back into a single-person motion.
In all cases, motions are generated continuously under incremental text updates.
The model produces smooth transitions between different scenarios within a single framework, while maintaining temporal coherence and spatial consistency.

\subsection{Motion Streaming: Quantitative Evaluation}
\begin{table}[tb]
  \caption{Quantitative comparison on continuous multi-segment streaming generation.}
  \label{table:eval_transition}
  \centering
  \resizebox{\linewidth}{!}{
  \begin{tabular}{l c c c c c c}
    \toprule
    \multirow{3}{*}{\bf Method} & \multicolumn{2}{c}{\bf Solo $\leftrightarrow$ Interaction} & \multicolumn{4}{c}{\bf Interaction $\leftrightarrow$ Interaction} \\
     & \multicolumn{2}{c}{Transition} & \multicolumn{2}{c}{Subsequence} & \multicolumn{2}{c}{Transition} \\
     &
    PJ \(\rightarrow\) &
    AUJ\(\downarrow\) &
    R@Top3\(\uparrow\) &
    FID\(\downarrow\) &
    PJ \(\rightarrow\) &
    AUJ\(\downarrow\) \\
    \midrule
    Ground truth & 0.046$^{\pm.000}$ & 0.672$^{\pm.000}$ & 0.643 $^{\pm.015}$ & 8.344$^{\pm.524}$ &  0.074$^{\pm.000}$ & 0.584$^{\pm.000}$ \\
    \midrule
    InterMask (Inpainting)~\cite{javedintermask} & -- & -- & {\bf0.631}$^{\pm.012}$ & {\bf28.554}$^{\pm.329}$&  0.269$^{\pm.002}$ & 22.562$^{\pm.057}$  \\
    InterMask (Adapted)~\cite{javedintermask} & 3.131$^{\pm.046}$ & 7.444$^{\pm.093}$ & 0.557$^{\pm.017}$ & 32.307$^{\pm.562}$ & 0.265$^{\pm.003}$ & 22.440$^{\pm.043}$ \\
    Ours (Streaming generation) & {\bf0.077}$^{\pm.001}$  & {\bf1.070}$^{\pm.005}$ & 0.509$^{\pm.016}$ & 30.620$^{\pm.561}$ & {\bf0.071}$^{\pm.001}$ & {\bf2.925}$^{\pm.010}$  \\
    \bottomrule
  \end{tabular}
  }
\end{table}

We adopt the jerk-based metrics introduced in FlowMDM~\cite{barquero2024seamless}.
Jerk is defined as the time derivative of acceleration, which is sensitive to abrupt motion changes.
Peak Jerk (PJ) measures the maximum instantaneous jerk magnitude across all joints, capturing extreme discontinuities, while Area Under the Jerk (AUJ) accumulates deviations from the average jerk over time, reflecting persistent smoothness irregularities.
For transition evaluation, PJ and AUJ are computed within a 2-second temporal window centered at the behavior transition point.
We evaluate two long-horizon streaming scenarios: 
(1) continuous interaction generation and 
(2) transitions between solo motion and interaction. 
Since no standardized benchmark exists for these tasks, we construct evaluation sequences from the InterHuman and HumanML3D test sets.

For interaction streaming, we randomly sample 64 groups of interaction descriptions and generate sequences composed of eight consecutive interaction segments using streaming inference. 
To analyze both transition quality and semantic fidelity, we report jerk-based transition metrics together with standard motion generation metrics computed on the individual interaction subsequences.
For comparison, we adapt the state-of-the-art method InterMask~\cite{javedintermask} to support history-conditioned generation by feeding previously generated frames as unmasked prefix tokens. 
As shown in Table~\ref{table:eval_transition}, InterMask with inpainting achieves the best R@Top3 and FID on the interaction subsequences.
However, we observe that this behavior arises because the model effectively starts a new motion conditioned on the next prompt rather than generating a smooth transition from the preceding motion.
In contrast, our streaming model explicitly enforces temporal continuity across segments.
As a result, the generated motion gradually adapts to satisfy the new textual condition, which slightly reduces semantic alignment on the isolated subsequences but produces substantially smoother long-horizon motion.
In addition, the initial interaction state of each sequence in our generation is conditioned on the previously generated sequence, whereas InterMask initializes each segment from a newly generated starting pose that may exhibit foot sliding.
These results show that our method significantly reduces motion discontinuities compared with the InterMask baseline, achieving lower PJ (0.071 vs.~0.265) and AUJ (2.925 vs.~22.440).

For solo--interaction transitions, we randomly construct 64 sequences consisting of a single-person motion, followed by an interaction motion, and another single-person motion.
We adapt and retrain InterMask on HumanML3D and InterHuman datasets to support both solo and duo generation by masking the second agent before interaction onset and activating it during the duo phase.
As shown in Table~\ref{table:eval_transition}, the adapted InterMask baseline exhibits large discontinuities at the solo-to-interaction boundary, with PJ/AUJ of 3.131/7.444, while ARMS remains close to ground truth with 0.077/1.070.
This gap reflects the difficulty of abruptly synthesizing a globally consistent two-person configuration after suppressing the second agent, whereas ARMS keeps the Anchor stream causally continuous and introduces the second agent through the Relational branch.

\subsection{Ablation Study}
\label{sec:exp-ablations}
\begin{table}[tb]
  \caption{Ablations on model design and inference hyperparameters.}
  \label{table:ablation}
  \centering
  \begin{tabular}{l c c c c }
    \toprule
    {\bf Methods} &
    \multicolumn{2}{c}{\bf Interaction generation} &
    \multicolumn{2}{c}{\bf Transition}  \\
    & {\bf R$@$Top3} \(\uparrow\) &
    {\bf FID}\(\downarrow\) &
    {\bf PJ} \(\rightarrow\) &
    {\bf AUJ}\(\downarrow\) \\
    \midrule
    Ground truth &  0.701$^{\pm.008}$ &
    0.273$^{\pm.007}$ & 0.074$^{\pm.000}$ & 0.584$^{\pm.000}$ \\
    \midrule
    w/ InterGen representation  
    & 0.723$^{\pm.004}$ & 4.553$^{\pm.065}$ & 0.130$^{\pm.001}$ & 3.367$^{\pm.022}$ \\
    w/o autoregressive & {\bf0.764}$^{\pm.004}$ & {\bf4.436}$^{\pm.069}$ & 0.114$^{\pm.006}$ & 3.209$^{\pm.024}$ \\
    \midrule
    Segment size $S = 1$ & 0.674$^{\pm.005}$ & 6.832$^{\pm0.115}$ & 0.131$^{\pm.003}$ & 3.353$^{\pm.079}$ \\
    Segment size $S = 3$ & 0.718$^{\pm.005}$ & 4.445$^{\pm.052}$ & 0.070$^{\pm.001}$ & {\bf2.708$^{\pm.011}$} \\
    Segment size $S = 10$ & 0.729$^{\pm.005}$ & 4.600$^{\pm.064}$ & 0.082$^{\pm.001}$ & 2.980$^{\pm.008}$ \\
    \midrule
    Latent dim $d = 32$ & 0.707$^{\pm.004}$ & 4.407$^{\pm.080}$ & 0.115$^{\pm.001}$ & 2.894$^{\pm.015}$ \\
    Latent dim $d = 128$ & 0.695$^{\pm.005}$ & 4.969$^{\pm.082}$ & 0.076$^{\pm.001}$ & 2.873$^{\pm.016}$ \\
    \midrule
    Sampling timesteps $K = 10$  & 0.722$^{\pm.005}$ & 4.731$^{\pm.068}$ & 0.105$^{\pm.001}$ & 3.064$^{\pm.008}$ \\
    Sampling timesteps $K = 20$  & 0.719$^{\pm.005}$ & 4.500$^{\pm.073}$ & 0.078$^{\pm.001}$ & 2.975$^{\pm.009}$ \\
    Refinement offset $\delta=1$ & 0.728$^{\pm.005}$ & 4.537$^{\pm.079}$ & 0.082$^{\pm.001}$ & 2.890$^{\pm.010}$ \\
    Refinement offset $\delta=3$ & 0.726$^{\pm.005}$ & 4.486$^{\pm.074}$ & {\bf0.075}$^{\pm.001}$ & 2.919$^{\pm.011}$ \\
    Refinement offset $\delta=10$ & 0.730$^{\pm.005}$ & 4.599$^{\pm.064}$ & 0.069$^{\pm.001}$ & 2.968$^{\pm.010}$ \\
    Refinement offset $\delta=50$ & 0.662$^{\pm.004}$ & 8.645$^{\pm.110}$ & 0.065$^{\pm.001}$ & 3.068$^{\pm.013}$ \\
    \midrule
    Ours & 0.723$^{\pm.008}$ & 4.446$^{\pm.116}$ & 0.071$^{\pm.001}$ & 2.925$^{\pm.010}$  \\
    \bottomrule
  \end{tabular}
\end{table}

We conduct ablation studies to analyze the contribution of key design components and the sensitivity to critical hyperparameters.
Results are summarized in Table~\ref{table:ablation}.
Replacing our Anchor--Relational representation with InterGen's non-canonical formulation~\cite{liang2024intergen} slightly improves interaction alignment, but significantly degrades transition smoothness, confirming that globally anchored modeling weakens long-horizon stability.
Removing the autoregressive streaming mechanism achieves the best single-window interaction metrics, yet increases transition jerk.
Setting segment size $S=1$, i.e., disabling joint latent refinement within segments, substantially degrades interaction realism and increases jerk, indicating insufficient temporal coupling.
Moderate segment sizes achieve the best balance, whereas larger segments slightly reduce transition smoothness, suggesting that coarse refinement weakens local adaptability.
Varying latent dimension $d$ shows that smaller latent spaces (e.g., $d=32$) improve interaction alignment but amplify jerk.
Increasing diffusion sampling steps $K$ improves transition stability and realism, while a moderate inter-segment offset achieves the lowest jerk, validating the effectiveness of the temporal refinement schedule.
Finally, the refinement offset $\delta$, which controls the temporal delay between denoising adjacent segments during streaming inference, achieves the best trade-off between interaction quality and transition smoothness at a moderate value. 

\section{Conclusion}
In this work, we present ARMS, an Anchor--Relational causal diffusion framework for streaming solo--social motion generation.
ARMS unifies solo motion and human--human interaction within a single incremental generative process through dynamics-asymmetric motion modeling, decoupling temporal progression from inter-person relational alignment.
Our causal relational diffusion model supports temporal refinement, variable-length history conditioning, and seamless transitions across one-/two-person social configurations.
Experiments demonstrate strong generation performance and substantially improved transition smoothness and social coherence in long-horizon streaming, suggesting a scalable pathway toward more flexible and realistic socially aware motion synthesis.

\noindent\textbf{Limitations.}
Although ARMS unifies solo and human--human interaction generation within a single streaming framework, several limitations remain.
First, the current formulation does not explicitly condition on surrounding agents or environmental context during solo motion generation, so crowd-level and scene-level collision avoidance are not directly addressed.
Second, this work focuses on up to two-person interaction scenarios and assumes that social mode is activated within the interaction-distance range covered by the training data.
If agents begin far outside this range, solo locomotion should first bring them into a plausible interaction range before activating social coupling.
While the relational formulation can be extended to variable numbers of agents through pairwise modeling~\cite{ota2025pino, liu2026hint}, scaling to multi-agent settings requires explicit reasoning over higher-order social structures.
Third, ARMS currently uses hard mode-aware relational gating to preserve strict causal consistency between solo and interaction modes.
While effective, this binary switching may still introduce abrupt changes in challenging transition cases.
Gradually blending relational attention across mode switches could further improve transition smoothness.

\section*{Acknowledgements}
This work was done during an internship at LY Corporation and was also supported by JST BOOST JPMJBS2423. 

\clearpage
\bibliographystyle{splncs04}
\bibliography{main}

@String(CVPR  = {IEEE Conf. Comput. Vis. Pattern Recog.})

@String(TOG   = {ACM Trans. Graph.})

@String(CVPR  = {CVPR})

@String(TOG   = {ACM TOG})

@inproceedings{barquero2024seamless,
  title={Seamless human motion composition with blended positional encodings},
  author={Barquero, German and Escalera, Sergio and Palmero, Cristina},
  booktitle={Proceedings of the IEEE/CVF Conference on Computer Vision and Pattern Recognition},
  pages={457--469},
  year={2024}
}

@inproceedings{petrovich2024multi,
  title={Multi-track timeline control for text-driven 3d human motion generation},
  author={Petrovich, Mathis and Litany, Or and Iqbal, Umar and Black, Michael J and Varol, Gul and Bin Peng, Xue and Rempe, Davis},
  booktitle={Proceedings of the IEEE/CVF Conference on Computer Vision and Pattern Recognition},
  pages={1911--1921},
  year={2024}
}

@inproceedings{zhuo2025infinidreamer,
  title={Infinidreamer: Arbitrarily long human motion generation via segment score distillation},
  author={Zhuo, Wenjie and Ma, Fan and Fan, Hehe},
  booktitle={Proceedings of the IEEE/CVF International Conference on Computer Vision},
  pages={14688--14698},
  year={2025}
}

@inproceedings{zhaodartcontrol,
  title={DartControl: A Diffusion-Based Autoregressive Motion Model for Real-Time Text-Driven Motion Control},
  author={Zhao, Kaifeng and Li, Gen and Tang, Siyu},
  booktitle={The Thirteenth International Conference on Learning Representations},
  year={2025}
}

@inproceedings{xiao2025motionstreamer,
  title={Motionstreamer: Streaming motion generation via diffusion-based autoregressive model in causal latent space},
  author={Xiao, Lixing and Lu, Shunlin and Pi, Huaijin and Fan, Ke and Pan, Liang and Zhou, Yueer and Feng, Ziyong and Zhou, Xiaowei and Peng, Sida and Wang, Jingbo},
  booktitle={Proceedings of the IEEE/CVF International Conference on Computer Vision},
  pages={10086--10096},
  year={2025}
}

@article{sui2026survey,
  title={A survey on human interaction motion generation},
  author={Sui, Kewei and Ghosh, Anindita and Hwang, Inwoo and Zhou, Bing and Wang, Jian and Guo, Chuan},
  journal={International Journal of Computer Vision},
  volume={134},
  number={3},
  pages={113},
  year={2026},
  publisher={Springer}
}

@article{liang2024intergen,
  title={Intergen: Diffusion-based multi-human motion generation under complex interactions},
  author={Liang, Han and Zhang, Wenqian and Li, Wenxuan and Yu, Jingyi and Xu, Lan},
  journal={International Journal of Computer Vision},
  volume={132},
  number={9},
  pages={3463--3483},
  year={2024},
  publisher={Springer}
}

@inproceedings{cenready,
  title={Ready-to-React: Online Reaction Policy for Two-Character Interaction Generation},
  author={Cen, Zhi and Pi, Huaijin and Peng, Sida and Shuai, Qing and Shen, Yujun and Bao, Hujun and Zhou, Xiaowei and Hu, Ruizhen},
  booktitle={The Thirteenth International Conference on Learning Representations},
  year={2025}
}

@inproceedings{chensitcom,
  title={Sitcom-Crafter: A Plot-Driven Human Motion Generation System in 3D Scenes},
  author={Chen, Jianqi and Hu, Panwen and Chang, Xiaojun and Shi, Zhenwei and Kampffmeyer, Michael and Liang, Xiaodan},
  booktitle={The Thirteenth International Conference on Learning Representations},
  year={2025}
}

@inproceedings{lim2025event,
  title={Event-Driven Storytelling with Multiple Lifelike Humans in a 3D Scene},
  author={Lim, Donggeun and Bae, Jinseok and Hwang, Inwoo and Lee, Seungmin and Lee, Hwanhee and Kim, Young Min},
  booktitle={Proceedings of the IEEE/CVF International Conference on Computer Vision},
  pages={11654--11664},
  year={2025}
}

@inproceedings{tevetclosd,
  title={CLoSD: Closing the Loop between Simulation and Diffusion for multi-task character control},
  author={Tevet, Guy and Raab, Sigal and Cohan, Setareh and Reda, Daniele and Luo, Zhengyi and Peng, Xue Bin and Bermano, Amit Haim and van de Panne, Michiel},
  booktitle={The Thirteenth International Conference on Learning Representations},
  year={2025}
}

@inproceedings{javedintermask,
  title={InterMask: 3D Human Interaction Generation via Collaborative Masked Modeling},
  author={Javed, Muhammad Gohar and Li, Xingyu and others},
  booktitle={The Thirteenth International Conference on Learning Representations},
  year={2025}
}

@inproceedings{petrovich2021action,
  title={Action-conditioned 3d human motion synthesis with transformer vae},
  author={Petrovich, Mathis and Black, Michael J and Varol, G{\"u}l},
  booktitle={Proceedings of the IEEE/CVF international conference on computer vision},
  pages={10985--10995},
  year={2021}
}

@inproceedings{guo2020action2motion,
  title={Action2motion: Conditioned generation of 3d human motions},
  author={Guo, Chuan and Zuo, Xinxin and Wang, Sen and Zou, Shihao and Sun, Qingyao and Deng, Annan and Gong, Minglun and Cheng, Li},
  booktitle={Proceedings of the 28th ACM international conference on multimedia},
  pages={2021--2029},
  year={2020}
}

@inproceedings{guo2022generating,
  title={Generating diverse and natural 3d human motions from text},
  author={Guo, Chuan and Zou, Shihao and Zuo, Xinxin and Wang, Sen and Ji, Wei and Li, Xingyu and Cheng, Li},
  booktitle={Proceedings of the IEEE/CVF conference on computer vision and pattern recognition},
  pages={5152--5161},
  year={2022}
}

@inproceedings{tevethuman,
  title={Human Motion Diffusion Model},
  author={Tevet, Guy and Raab, Sigal and Gordon, Brian and Shafir, Yoni and Cohen-Or, Daniel and Bermano, Amit Haim},
  booktitle={The Eleventh International Conference on Learning Representations},
  year={2023},
}

@inproceedings{ghosh2025duetgen,
  title={Duetgen: Music driven two-person dance generation via hierarchical masked modeling},
  author={Ghosh, Anindita and Zhou, Bing and Dabral, Rishabh and Wang, Jian and Golyanik, Vladislav and Theobalt, Christian and Slusallek, Philipp and Guo, Chuan},
  booktitle={Proceedings of the Special Interest Group on Computer Graphics and Interactive Techniques Conference Conference Papers},
  pages={1--11},
  year={2025}
}

@article{alexanderson2023listen,
  title={Listen, denoise, action! audio-driven motion synthesis with diffusion models},
  author={Alexanderson, Simon and Nagy, Rajmund and Beskow, Jonas and Henter, Gustav Eje},
  journal={ACM Transactions on Graphics (TOG)},
  volume={42},
  number={4},
  pages={1--20},
  year={2023},
  publisher={ACM New York, NY, USA}
}

@inproceedings{liu2024emage,
  title={Emage: Towards unified holistic co-speech gesture generation via expressive masked audio gesture modeling},
  author={Liu, Haiyang and Zhu, Zihao and Becherini, Giorgio and Peng, Yichen and Su, Mingyang and Zhou, You and Zhe, Xuefei and Iwamoto, Naoya and Zheng, Bo and Black, Michael J},
  booktitle={Proceedings of the IEEE/CVF conference on computer vision and pattern recognition},
  pages={1144--1154},
  year={2024}
}

@inproceedings{mullen2023placing,
  title={Placing human animations into 3d scenes by learning interaction-and geometry-driven keyframes},
  author={Mullen, James F and Kothandaraman, Divya and Bera, Aniket and Manocha, Dinesh},
  booktitle={Proceedings of the IEEE/CVF Winter Conference on Applications of Computer Vision},
  pages={300--310},
  year={2023}
}

@inproceedings{zhao2023synthesizing,
  title={Synthesizing diverse human motions in 3d indoor scenes},
  author={Zhao, Kaifeng and Zhang, Yan and Wang, Shaofei and Beeler, Thabo and Tang, Siyu},
  booktitle={Proceedings of the IEEE/CVF international conference on computer vision},
  pages={14738--14749},
  year={2023}
}

@article{ouyang2025motion,
  title={Motion-r1: Chain-of-thought reasoning and reinforcement learning for human motion generation},
  author={Ouyang, Runqi and Li, Haoyun and Zhang, Zhenyuan and Wang, Xiaofeng and Zhu, Zheng and Huang, Guan and Wang, Xingang},
  journal={arXiv e-prints},
  pages={arXiv--2506},
  year={2025}
}

@inproceedings{shafirhuman,
  title={Human Motion Diffusion as a Generative Prior},
  author={Shafir, Yoni and Tevet, Guy and Kapon, Roy and Bermano, Amit Haim},
  booktitle={The Twelfth International Conference on Learning Representations},
  year={2024}
}

@inproceedings{tanaka2023role,
  title={Role-aware interaction generation from textual description},
  author={Tanaka, Mikihiro and Fujiwara, Kent},
  booktitle={Proceedings of the IEEE/CVF international conference on computer vision},
  pages={15999--16009},
  year={2023}
}

@inproceedings{ruiz2024in2in,
  title={in2in: Leveraging individual information to generate human interactions},
  author={Ruiz-Ponce, Pablo and Barquero, German and Palmero, Cristina and Escalera, Sergio and Garc{\'\i}a-Rodr{\'\i}guez, Jos{\'e}},
  booktitle={Proceedings of the IEEE/CVF Conference on Computer Vision and Pattern Recognition},
  pages={1941--1951},
  year={2024}
}

@article{wu2025text2interact,
  title={Text2interact: High-fidelity and diverse text-to-two-person interaction generation},
  author={Wu, Qingxuan and Dou, Zhiyang and Guo, Chuan and Huang, Yiming and Feng, Qiao and Zhou, Bing and Wang, Jian and Liu, Lingjie},
  journal={arXiv preprint arXiv:2510.06504},
  year={2025}
}

@inproceedings{ruiz2025mixermdm,
  title={Mixermdm: Learnable composition of human motion diffusion models},
  author={Ruiz-Ponce, Pablo and Barquero, German and Palmero, Cristina and Escalera, Sergio and Garc{\'\i}a-Rodr{\'\i}guez, Jos{\'e}},
  booktitle={Proceedings of the Computer Vision and Pattern Recognition Conference},
  pages={12380--12390},
  year={2025}
}

@inproceedings{yusocialgen,
  title={SocialGen: Modeling Multi-Human Social Interaction with Language Models},
  author={Yu, Heng and Zhang, Juze and Chen, Changan and Xiang, Tiange and Fang, Yusu and Niebles, Juan Carlos and Adeli, Ehsan},
  booktitle={Thirteenth International Conference on 3D Vision},
  year={2025}
}

@inproceedings{wang2025timotion,
  title={Timotion: Temporal and interactive framework for efficient human-human motion generation},
  author={Wang, Yabiao and Wang, Shuo and Zhang, Jiangning and Fan, Ke and Wu, Jiafu and Xue, Zhucun and Liu, Yong},
  booktitle={Proceedings of the Computer Vision and Pattern Recognition Conference},
  pages={7169--7178},
  year={2025}
}

@inproceedings{chen2024taming,
  title={Taming diffusion probabilistic models for character control},
  author={Chen, Rui and Shi, Mingyi and Huang, Shaoli and Tan, Ping and Komura, Taku and Chen, Xuelin},
  booktitle={ACM SIGGRAPH 2024 Conference Papers},
  pages={1--10},
  year={2024}
}

@article{shi2024interactive,
  title={Interactive character control with auto-regressive motion diffusion models},
  author={Shi, Yi and Wang, Jingbo and Jiang, Xuekun and Lin, Bingkun and Dai, Bo and Peng, Xue Bin},
  journal={ACM Transactions on Graphics (TOG)},
  volume={43},
  number={4},
  pages={1--14},
  year={2024},
  publisher={ACM New York, NY, USA}
}

@inproceedings{zhang2025primal,
  title={Primal: Physically reactive and interactive motor model for avatar learning},
  author={Zhang, Yan and Feng, Yao and Cseke, Alp{\'a}r and Saini, Nitin and Bajandas, Nathan and Heron, Nicolas and Black, Michael J},
  booktitle={Proceedings of the IEEE/CVF International Conference on Computer Vision},
  pages={12725--12736},
  year={2025}
}

@article{ruiz2025interact2ar,
  title={Interact2Ar: Full-Body Human-Human Interaction Generation via Autoregressive Diffusion Models},
  author={Ruiz-Ponce, Pablo and Escalera, Sergio and Garc{\'\i}a-Rodr{\'\i}guez, Jos{\'e} and Deng, Jiankang and Potamias, Rolandos Alexandros},
  journal={arXiv preprint arXiv:2512.19692},
  year={2025}
}

@article{liu2026hint,
  title={HINT: Hierarchical Interaction Modeling for Autoregressive Multi-Human Motion Generation},
  author={Liu, Mengge and Di, Yan and Wang, Gu and Qu, Yun and Zhu, Dekai and Li, Yanyan and Ji, Xiangyang},
  journal={arXiv preprint arXiv:2601.20383},
  year={2026}
}

@inproceedings{zhou2019continuity,
  title={On the continuity of rotation representations in neural networks},
  author={Zhou, Yi and Barnes, Connelly and Lu, Jingwan and Yang, Jimei and Li, Hao},
  booktitle={Proceedings of the IEEE/CVF conference on computer vision and pattern recognition},
  pages={5745--5753},
  year={2019}
}

@inproceedings{xu2024inter,
  title={Inter-x: Towards versatile human-human interaction analysis},
  author={Xu, Liang and Lv, Xintao and Yan, Yichao and Jin, Xin and Wu, Shuwen and Xu, Congsheng and Liu, Yifan and Zhou, Yizhou and Rao, Fengyun and Sheng, Xingdong and others},
  booktitle={Proceedings of the IEEE/CVF conference on computer vision and pattern recognition},
  pages={22260--22271},
  year={2024}
}

@inproceedings{mahmood2019amass,
  title={AMASS: Archive of motion capture as surface shapes},
  author={Mahmood, Naureen and Ghorbani, Nima and Troje, Nikolaus F and Pons-Moll, Gerard and Black, Michael J},
  booktitle={Proceedings of the IEEE/CVF international conference on computer vision},
  pages={5442--5451},
  year={2019}
  }

@inproceedings{pavlakos2019expressive,
  title={Expressive body capture: 3d hands, face, and body from a single image},
  author={Pavlakos, Georgios and Choutas, Vasileios and Ghorbani, Nima and Bolkart, Timo and Osman, Ahmed AA and Tzionas, Dimitrios and Black, Michael J},
  booktitle={Proceedings of the IEEE/CVF conference on computer vision and pattern recognition},
  pages={10975--10985},
  year={2019}
}

@article{chen2021choreomaster,
  title={Choreomaster: choreography-oriented music-driven dance synthesis},
  author={Chen, Kang and Tan, Zhipeng and Lei, Jin and Zhang, Song-Hai and Guo, Yuan-Chen and Zhang, Weidong and Hu, Shi-Min},
  journal={ACM Transactions on Graphics (TOG)},
  volume={40},
  number={4},
  pages={1--13},
  year={2021},
  publisher={ACM New York, NY, USA}
}

@inproceedings{li2021ai,
  title={Ai choreographer: Music conditioned 3d dance generation with aist++},
  author={Li, Ruilong and Yang, Shan and Ross, David A and Kanazawa, Angjoo},
  booktitle={Proceedings of the IEEE/CVF international conference on computer vision},
  pages={13401--13412},
  year={2021}
}

@article{yu2026causal,
  title={Causal Motion Diffusion Models for Autoregressive Motion Generation},
  author={Yu, Qing and Watanabe, Akihisa and Fujiwara, Kent},
  journal={arXiv preprint arXiv:2602.22594},
  year={2026}
}

@article{chen2024diffusion,
  title={Diffusion forcing: Next-token prediction meets full-sequence diffusion},
  author={Chen, Boyuan and Mart{\'\i} Mons{\'o}, Diego and Du, Yilun and Simchowitz, Max and Tedrake, Russ and Sitzmann, Vincent},
  journal={Advances in Neural Information Processing Systems},
  volume={37},
  pages={24081--24125},
  year={2024}
}

@inproceedings{peebles2023scalable,
  title={Scalable diffusion models with transformers},
  author={Peebles, William and Xie, Saining},
  booktitle={Proceedings of the IEEE/CVF international conference on computer vision},
  pages={4195--4205},
  year={2023}
}

@article{su2024roformer,
  title={Roformer: Enhanced transformer with rotary position embedding},
  author={Su, Jianlin and Ahmed, Murtadha and Lu, Yu and Pan, Shengfeng and Bo, Wen and Liu, Yunfeng},
  journal={Neurocomputing},
  volume={568},
  pages={127063},
  year={2024},
  publisher={Elsevier}
}

@article{sanh2019distilbert,
  title={DistilBERT, a distilled version of BERT: smaller, faster, cheaper and lighter},
  author={Sanh, Victor and Debut, Lysandre and Chaumond, Julien and Wolf, Thomas},
  journal={arXiv preprint arXiv:1910.01108},
  year={2019}
}

@inproceedings{ota2025pino,
  title={Pino: Person-interaction noise optimization for long-duration and customizable motion generation of arbitrary-sized groups},
  author={Ota, Sakuya and Yu, Qing and Fujiwara, Kent and Ikehata, Satoshi and Sato, Ikuro},
  booktitle={Proceedings of the IEEE/CVF International Conference on Computer Vision},
  pages={10676--10685},
  year={2025}
}

@inproceedings{cai2024digital,
  title={Digital life project: Autonomous 3d characters with social intelligence},
  author={Cai, Zhongang and Jiang, Jianping and Qing, Zhongfei and Guo, Xinying and Zhang, Mingyuan and Lin, Zhengyu and Mei, Haiyi and Wei, Chen and Wang, Ruisi and Yin, Wanqi and others},
  booktitle={Proceedings of the IEEE/CVF conference on computer vision and pattern recognition},
  pages={582--592},
  year={2024}
}

@inproceedings{dabral2023mofusion,
  title={Mofusion: A framework for denoising-diffusion-based motion synthesis},
  author={Dabral, Rishabh and Mughal, Muhammad Hamza and Golyanik, Vladislav and Theobalt, Christian},
  booktitle={Proceedings of the IEEE/CVF conference on computer vision and pattern recognition},
  pages={9760--9770},
  year={2023}
}

@inproceedings{guo2022tm2t,
  title={Tm2t: Stochastic and tokenized modeling for the reciprocal generation of 3d human motions and texts},
  author={Guo, Chuan and Zuo, Xinxin and Wang, Sen and Cheng, Li},
  booktitle={European Conference on Computer Vision},
  pages={580--597},
  year={2022},
  organization={Springer}
}

@article{jiang2023motiongpt,
  title={Motiongpt: Human motion as a foreign language},
  author={Jiang, Biao and Chen, Xin and Liu, Wen and Yu, Jingyi and Yu, Gang and Chen, Tao},
  journal={Advances in Neural Information Processing Systems},
  volume={36},
  pages={20067--20079},
  year={2023}
}

@article{zhu2025motiongpt3,
  title={Motiongpt3: Human motion as a second modality},
  author={Zhu, Bingfan and Jiang, Biao and Wang, Sunyi and Tang, Shixiang and Chen, Tao and Luo, Linjie and Zheng, Youyi and Chen, Xin},
  journal={arXiv preprint arXiv:2506.24086},
  year={2025}
}

@inproceedings{petrovich2022temos,
  title={Temos: Generating diverse human motions from textual descriptions},
  author={Petrovich, Mathis and Black, Michael J and Varol, G{\"u}l},
  booktitle={European conference on computer vision},
  pages={480--497},
  year={2022},
  organization={Springer}
}

@article{tevet2022human,
  title={Human motion diffusion model},
  author={Tevet, Guy and Raab, Sigal and Gordon, Brian and Shafir, Yonatan and Cohen-Or, Daniel and Bermano, Amit H},
  journal={arXiv preprint arXiv:2209.14916},
  year={2022}
}

@article{sahili2025text,
  title={Text-driven motion generation: Overview, challenges and directions},
  author={Sahili, Ali Rida and Neji, Najett and Tabia, Hedi},
  journal={arXiv preprint arXiv:2505.09379},
  year={2025}
}

@article{meng2025absolute,
  title={Absolute coordinates make motion generation easy},
  author={Meng, Zichong and Han, Zeyu and Peng, Xiaogang and Xie, Yiming and Jiang, Huaizu},
  journal={arXiv preprint arXiv:2505.19377},
  year={2025}
}

@inproceedings{albergobuilding,
  title={Building Normalizing Flows with Stochastic Interpolants},
  author={Albergo, Michael Samuel and Vanden-Eijnden, Eric},
  booktitle={The Eleventh International Conference on Learning Representations},
  year={2023}
}

@article{albergo2025stochastic,
  title={Stochastic interpolants: A unifying framework for flows and diffusions},
  author={Albergo, Michael and Boffi, Nicholas M and Vanden-Eijnden, Eric},
  journal={Journal of Machine Learning Research},
  volume={26},
  number={209},
  pages={1--80},
  year={2025}
}

@inproceedings{lugmayr2023inpainting,
  title={Inpainting using denoising diffusion probabilistic models},
  author={Lugmayr, Andreas and Danelljan, Martin and Romero, Andres and Yu, Fisher and Timofte, Radu and Van Gool, L Repaint},
  booktitle={Proceedings of the IEEE/CVF conference on computer vision and pattern recognition},
  pages={11461--11471},
  year={2023}
}

@inproceedings{chen2023executing,
  title={Executing your commands via motion diffusion in latent space},
  author={Chen, Xin and Jiang, Biao and Liu, Wen and Huang, Zilong and Fu, Bin and Chen, Tao and Yu, Gang},
  booktitle={Proceedings of the IEEE/CVF conference on computer vision and pattern recognition},
  pages={18000--18010},
  year={2023}
}

@inproceedings{zhong2023attt2m,
  title={Attt2m: Text-driven human motion generation with multi-perspective attention mechanism},
  author={Zhong, Chongyang and Hu, Lei and Zhang, Zihao and Xia, Shihong},
  booktitle={Proceedings of the IEEE/CVF international conference on computer vision},
  pages={509--519},
  year={2023}
}

@inproceedings{guo2024momask,
  title={Momask: Generative masked modeling of 3d human motions},
  author={Guo, Chuan and Mu, Yuxuan and Javed, Muhammad Gohar and Wang, Sen and Cheng, Li},
  booktitle={Proceedings of the IEEE/CVF Conference on Computer Vision and Pattern Recognition},
  pages={1900--1910},
  year={2024}
}

@inproceedings{zhang2023generating,
  title={T2M-GPT: Generating Human Motion from Textual Descriptions with Discrete Representations},
  author={Zhang, Jianrong and Zhang, Yangsong and Cun, Xiaodong and Huang, Shaoli and Zhang, Yong and Zhao, Hongwei and Lu, Hongtao and Shen, Xi},
  booktitle={Proceedings of the IEEE/CVF Conference on Computer Vision and Pattern Recognition (CVPR)},
  year={2023},
}

\title{ARMS: Anchor--Relational Motion Streaming for Seamless Solo-Social Motion Transitions\\
Supplementary Material}
\titlerunning{ARMS: Anchor-Relational Motion Streaming}

\author{Huakun Liu\inst{1}\orcidlink{0000-0002-9130-2519} \and
Qing Yu\inst{2}\orcidlink{0000-0001-6965-9581} \and
Kent Fujiwara\inst{2}\orcidlink{0000-0002-2205-6115} \and
Hideaki Uchiyama\inst{1}\orcidlink{0000-0002-6119-1184} \and
Kiyoshi Kiyokawa\inst{1}\orcidlink{0000-0003-2260-1707}}

\authorrunning{H.~Liu et al.}

\institute{Nara Institute of Science and Technology, Ikoma, Japan \and
LY Corporation, Tokyo, Japan \\
\email{\{liu.huakun.li0,hideaki.uchiyama,kiyo\}@is.naist.jp, \{yu.qing,kent.fujiwara\}@lycorp.co.jp}}

\maketitle

\renewcommand{\thesection}{\Alph{section}}
\setcounter{section}{0}
\section{Implementation Details}
\subsection{Training Details}
\subsubsection{Temporal Autoencoders} 
We train a causal temporal variational autoencoder to compress motion sequences into a continuous latent space.
All datasets are converted to our anchor-relational motion representation.
Specifically, we extract the yaw angle from the root orientation and the $xz$ translation from the root trajectory while keeping the remaining joint features unchanged.
The encoder and decoder are implemented using causal 1D convolution and residual blocks.
The encoder progressively downsamples the temporal dimension by a factor of $4$, producing latent tokens with dimension $d=64$.
The model is trained on motion windows of $64$ frames with a stride of $10$ frames using the AdamW optimizer with a learning rate of $2\times10^{-4}$ and $\beta=(0.9,0.99)$.
We use a batch size of $256$ and apply linear learning-rate warmup followed by step decay at $70\%$ and $85\%$ of the total training iterations.
The training objective consists of latent reconstruction loss, root reconstruction loss, and KL regularization, following prior work~\cite{xiao2025motionstreamer}.
Training is performed on a single NVIDIA~RTX~5090 GPU.
Training takes approximately one hour on InterHuman and around two hours when jointly training on HumanML3D and InterHuman.

\subsubsection{Causal Relational Diffusion}
The motion diffusion model is implemented as a causal relational transformer operating in the latent space produced by the temporal autoencoder.
We use a transformer with $8$ layers, hidden dimension $512$, feed-forward dimension $1024$, and $4$ attention heads.
Text conditions are encoded using a frozen DistilBERT encoder and projected to the transformer dimension.
During training, motion sequences are first encoded into temporally downsampled latent tokens using the pretrained causal autoencoder.
We train the diffusion model with a flow-matching formulation using velocity prediction.
Given a clean latent trajectory $\mathbf{z}_k$, we first construct a noised latent trajectory by perturbing each latent token with segment-wise noise,
\begin{equation}
    \tilde{\mathbf{z}}_{k} = \alpha_{s(k)}\,\mathbf{z}_{k} + (1-\alpha_{s(k)})\,\boldsymbol{\epsilon}_{k},
  \qquad \boldsymbol{\epsilon}_{k}\sim\mathcal{N}(\mathbf{0},\mathbf{I}),
  \label{eq:add_noise}
\end{equation}
where $\alpha_{s(k)}$ is determined by the noise level of the segment containing timestep $k$.
The transformer then predicts the target velocity field $\mathbf{u}_k$ along the linear transport path from the noised latent trajectory.
The training objective is the mean squared error between the predicted and target velocity fields,
\begin{equation}
  \mathcal{L}_{\mathrm{fm}}
  =
  \frac{1}{|\mathcal{M}|}
  \sum_{(b,k)\in\mathcal{M}}
  \mathrm{mean}_{d}
  \left[
    \left\|
      \hat{\mathbf{u}}_{b,k,d} - \mathbf{u}_{b,k,d}
    \right\|_2^2
  \right],
\end{equation}
where $\mathcal{M}$ denotes the set of valid non-padding latent timesteps and the mean is taken over the latent feature dimension.
In practice, we compute the squared error over the latent dimension for each latent timestep and average it over all valid latent tokens in the batch.
Training is performed with the AdamW optimizer using a learning rate of $2\times10^{-4}$, $\beta=(0.9,0.99)$, and weight decay $10^{-5}$.
We use a batch size of $64$ and apply linear learning-rate warmup followed by step decay at $50\%$, $70\%$, and $85\%$ of the total training iterations.
Classifier-free guidance is implemented by randomly dropping text conditions with probability $0.1$ during training.
Latent sequences are partitioned into fixed-size segments during training, with $5$ latent steps per segment in the latent timeline.
All experiments are trained on a single RTX~5090 GPU.
Training takes approximately $3$ hours on InterHuman and around $7$ hours on HumanML3D.

\subsection{Inference Details}
At inference time, motion is generated by progressively refining latent segments under a temporally structured noise schedule.
Given a latent state $\mathbf{z}^{(i)}$ at noise level $\alpha^{(i)}$, where $i\in \{1, \dots, K\}$ denotes the denoising step, the diffusion model predicts a velocity field that updates the latent according to
\begin{equation}
\mathbf{z}^{(i+1)} =
\mathbf{z}^{(i)} +
\big(\alpha^{(i+1)} - \alpha^{(i)}\big)
\,\mathbf{v}_\theta(\mathbf{z}^{(i)},\alpha^{(i)},\mathbf{c}),
\end{equation}
where $\mathbf{v}_\theta$ is the velocity predicted by the causal relational transformer conditioned on text embedding $\mathbf{c}$.
Classifier-free guidance is applied during inference by combining conditional and unconditional predictions with guidance scale $2.5$.

\subsubsection{Prompt-conditioned streaming generation}

For long-horizon generation under a fixed textual instruction, motion is generated using a segment-based sliding window in the latent space.
The latent timeline is divided into contiguous segments, each containing a fixed number of latent steps.
At each iteration, a new noisy segment is appended to the current sequence while previously generated segments remain within the context window.
To efficiently refine long sequences, we employ a staggered scheduling that assigns different noise levels to different segments~\cite{chen2024diffusion, yu2026causal}.
Earlier segments receive lower noise levels while later segments remain noisier.
This staggered scheduling allows multiple segments to be refined jointly during the same denoising process while preserving causal temporal ordering.
Consequently, earlier segments gradually converge to deterministic motion while later segments remain partially noisy and continue to be refined.
This design enables arbitrarily long motion sequences to be generated without restarting the diffusion process.
When textual instructions change during generation, the new prompt replaces the conditioning embedding for future segments while previously generated motion remains fixed as context.
To preserve temporal continuity, we adopt a resampling strategy used in RePaint~\cite{lugmayr2023inpainting}, where a short history of previously generated latent tokens is retained and injected into the denoising process.
Specifically, the model first predicts a clean latent estimate
\begin{equation}
\hat{\mathbf{z}}_0 =
\mathbf{z}^{(i)} +
(1-\alpha^{(i)})
\,\mathbf{v}_\theta(\mathbf{z}^{(i)},\alpha^{(i)},\mathbf{c}),
\end{equation}
after which the historical context tokens are restored at their corresponding temporal positions.
The latent sequence is then re-noised to the next scheduled noise level before continuing the denoising process.
The incremental generation process is summarized in~\cref{alg:full_seq} and~\cref{alg:transition}.

\begin{algorithm}[t]
\caption{Streaming generation with a fixed prompt}
\label{alg:full_seq}
\begin{algorithmic}[1]
\Require text prompt $\tau$, target per-agent length $L$ ($2L$ in total for interaction)
\State Encode $\tau$ into condition vector $\mathbf{c}$
\State Initialize latent sequence $\mathbf{Z} \leftarrow \emptyset$
\While{$|\mathbf{Z}| < L$}
\State Append new noisy segment $\mathbf{z}_{new} \sim \mathcal{N}(0,I)$
\State $\mathbf{Z} \leftarrow [\mathbf{Z},\mathbf{z}_{new}]$
\State Select sliding window $\mathbf{Z}_{win}$
\State Construct noise matrix $\Lambda$
\For{denoising step $i=1\dots K$}
\State Predict velocity $\mathbf{v}_\theta(\mathbf{Z}_{win},\Lambda_i,\mathbf{c})$
\State Update latents
\[
\mathbf{Z}_{win} \leftarrow
\mathbf{Z}_{win} +
(\Lambda_{i+1}-\Lambda_i)\mathbf{v}_\theta
\]
\EndFor
\EndWhile
\State \Return $\mathbf{Z}$
\end{algorithmic}
\end{algorithm}

\begin{algorithm}[t]
\caption{Streaming generation with prompt transition}
\label{alg:transition}
\begin{algorithmic}[1]
\Require context latents $\mathbf{H}$, new prompt $\tau'$
\State Encode $\tau'$ to condition vector $\mathbf{c}$
\State Initialize current window $\mathbf{Z}$ with noisy future segment
\State Construct noise matrix $\Lambda$
\For{denoising step $i=1\dots K$}
\State Predict velocity
\[
\mathbf{v} = \mathbf{v}_\theta(\mathbf{Z},\Lambda_i,\mathbf{c})
\]
\State Estimate clean latent
\[
\hat{\mathbf{Z}}_0 =
\mathbf{Z} + (1-\Lambda_i)\mathbf{v}
\]
\State Restore historical context
\[
\hat{\mathbf{Z}}_0[\text{context}] \leftarrow \mathbf{H}
\]
\State Sample noise $\boldsymbol{\epsilon} \sim \mathcal{N}(0,I)$
\State Re-noise to next level
\[
\mathbf{Z} \leftarrow
\Lambda_{i+1}\hat{\mathbf{Z}}_0 +
(1-\Lambda_{i+1})\boldsymbol{\epsilon}
\]
\EndFor
\State \Return refined latent sequence
\end{algorithmic}
\end{algorithm}

\subsubsection{Agent configuration transitions}
Our framework supports dynamic transitions between solo motion and human--human interaction during generation.
Let $\mathbf{z}^{(A)}$ and $\mathbf{z}^{(R)}$ denote the latent sequences of the anchor and relational agents, respectively.
During diffusion refinement, each active branch is updated by the velocity predictor
\begin{equation}
\mathbf{z}^{(j)}_{i+1}
=
\mathbf{z}^{(j)}_{i}
+
(\alpha_{i+1}-\alpha_i)\,
\mathbf{v}_\theta^{(j)}(\mathbf{z}^{(A)}_{i},\mathbf{z}^{(R)}_{i},\alpha_i,\mathbf{c}),
\qquad j\in\{A,R\},
\end{equation}
with cross-agent visibility controlled by the attention mask.
When a second agent appears, the relational branch is activated while the anchor branch is preserved from the previously generated solo motion.
The new relational latent can be initialized from noise,
\begin{equation}
\mathbf{z}^{(R)}_0 \sim \mathcal{N}(0,\mathbf{I}),
\end{equation}
or from an encoded motion context when such initialization is available.
Cross-agent attention is then enabled so that the relational branch can align with the historical anchor states $\mathbf{z}^{(A)}_{H}$:
\begin{equation}
\mathbf{v}_\theta^{(R)}
=
\mathbf{v}_\theta^{(R)}(\mathbf{z}^{(R)},\mathbf{z}^{(A)}, \mathbf{z}^{(A)}_H,\alpha,\mathbf{c}).
\end{equation}
This allows the newly introduced agent to align with the existing motion.
When the interaction ends, cross-agent attention is disabled and each branch can continue independently as a solo motion stream.
If the former anchor agent continues, its latent trajectory is refined using the same streaming generation process as~\cref{alg:transition}, conditioned on the existing anchor latent history.
Likewise, the former relational branch can be reassigned as the new anchor stream and continued using its own latent history under the same streaming generation procedure.
In both cases, generation proceeds under the solo-mode attention mask.
This role reassignment allows either participant in the interaction to continue as an individual motion stream without restarting generation.
The overall transition procedure is summarized in~\cref{alg:agent_transition}.

\begin{algorithm}[t]
\caption{Agent configuration transitions during streaming generation}
\label{alg:agent_transition}
\begin{algorithmic}[1]
\Require previous latents $\mathbf{z}^{(A)}$, $\mathbf{z}^{(R)}$, transition mode $m$

\If{solo $\rightarrow$ interaction}
    \State Preserve anchor latent $\mathbf{z}^{(A)}$
    \If{initialize new agent from noise}
        \State $\mathbf{z}^{(R)} \sim \mathcal{N}(0,I)$
    \Else
        \State $\mathbf{z}^{(R)} \leftarrow$ Re-encode relational dynamics from the available motion context
    \EndIf
    \State Enable cross-agent attention
    \For{denoising step $i=1\dots K$}
        \State Update both $\mathbf{z}^{(A)}$ and $\mathbf{z}^{(R)}$
    \EndFor
\EndIf

\If{interaction $\rightarrow$ solo}
    \State Disable cross-agent attention
    \State Select continuing branch $\mathbf{z}^{(\star)} \in \{\mathbf{z}^{(A)}, \mathbf{z}^{(R)}\}$
    \State Reassign anchor slot $\tilde{\mathbf{z}}^{(A)} \leftarrow \mathbf{z}^{(\star)}$
    \For{denoising step $i=1\dots K$}
        \State Update $\tilde{\mathbf{z}}^{(A)}$ under solo-mode masking
    \EndFor
\EndIf

\end{algorithmic}
\end{algorithm}

\subsection{Practical Usage and Scope}
ARMS assumes a shared text condition within one model instance.
When two agents need to perform simultaneous but independent solo motions before interacting, a practical usage is to run two independent ARMS streams in parallel, each with its own prompt, duration, and timing.
Once a shared interaction prompt begins, the streams can be merged by activating the relational branch and continuing generation under the interaction-mode mask.

The Anchor--Relational representation is trained on the interaction-distance distribution present in the data.
If agents start far outside this range, directly activating social mode can make the relational offset out-of-distribution.
In this case, the intended procedure is to first generate solo locomotion that brings the agents into a plausible interaction range and then activate social mode.

ARMS uses hard mode-aware relational gating to preserve strict causal consistency between solo and interaction modes.
For user control, the generation mode can be specified explicitly.
As an optional automatic alternative, we train a lightweight binary MLP classifier on frozen DistilBERT text embeddings to predict whether the prompt describes solo motion or two-person interaction.
The classifier mean-pools valid token embeddings and applies a MLP, achieving 99.9\% mode classification accuracy.

\subsection{On the InterX Dataset}
We additionally evaluated our method on the InterX dataset, a large-scale human--human interaction benchmark based on the SMPL-X representation~\cite{pavlakos2019expressive}.
InterX contains 11{,}388 motion sequences, each paired with three textual descriptions.
Compared with InterHuman, InterX uses the SMPL-X skeleton with 54 body, hand, and face joints plus the root joint, resulting in a significantly higher-dimensional motion representation.
To ensure compatibility with our model, we convert the InterX motions into the same anchor--relational representation used for training.
This conversion follows the same preprocessing procedure as for the HumanML3D and InterHuman datasets.

\section{More Experimental Results}

\subsection{On the HumanML3D Dataset}
\begin{table}[tb]
  \caption{\textbf{Quantitative evaluation} on the \textbf{HumanML3D (272-dim)} test sets.
  Values are reported with 95\% confidence intervals.
  The arrow $\rightarrow$ indicates that performance closer to ground truth is preferred.
  Best and second-best are \textbf{bold} and \underline{underlined}.}
  \label{table:humanml3d-eval-details}
  \centering
  \resizebox{\linewidth}{!}{
  \begin{tabular}{l c c c c c c c }
    \toprule
    \multirow{2.5}{*}{Method} &
    \multicolumn{3}{c}{R Precision\(\uparrow\)} &
    \multirow{2.5}{*}{FID\(\downarrow\)} &
    \multirow{2.5}{*}{MM Dist\(\downarrow\)} &
    \multirow{2.5}{*}{Diversity\(\rightarrow\)} \\
    \cmidrule{2-4}
    & Top 1 & Top 2 & Top 3 & & & & \\
    \midrule
    Ground Truth &
    0.002 &
    0.711&
    0.851&
    0.903&
    15.805&
    27.670\\
    \midrule
    MDM~\cite{tevethuman}&
    0.524&
    0.693&
    0.773&
    22.557&
    17.223&
    27.355\\
    MLD~\cite{chen2023executing}&
    0.548&
    0.732&
    0.805&
    17.226&
    16.338&
    26.551\\
    T2M-GPT~\cite{zhang2023generating}&
    0.608&
    0.772&
    0.831&
    11.175&
    16.810&
    \underline{27.617}\\
    MotionGPT~\cite{jiang2023motiongpt} &
    0.436&
    0.598&
    0.668&
    14.175&
    17.890&
    27.014\\
    MoMask~\cite{guo2024momask} &
    0.622&
    0.782&
    0.850&
    10.731&
    16.128&
    27.317\\
    AttT2M~\cite{zhong2023attt2m} &
    0.590&
    0.767&
    0.837&
    15.438&
    \underline{15.734}&
    26.680\\
    MotionStreamer~\cite{xiao2025motionstreamer} &
    \underline{0.631}&
    \underline{0.784}&
    \underline{0.851}&
    \underline{10.724}&
    16.639&
    {\bf 27.657}\\
    \midrule
    Ours (streaming generation)  &
    {\bf 0.769}&
    {\bf 0.916}&
    {\bf 0.952}&
    {\bf 4.975}&
    {\bf 14.663}&
    27.266\\
    \bottomrule
  \end{tabular}
  }
\end{table}

We evaluate our method on the HumanML3D dataset using the 272-dim motion representation proposed in MotionStreamer~\cite{xiao2025motionstreamer}.
This representation is convertible with the anchor branch of our representation and allows direct comparison with prior streaming text-to-motion models.
Quantitative results on the HumanML3D test set are reported in~\cref{table:humanml3d-eval-details}.
Our streaming generation model achieves the best performance across most evaluation metrics, significantly improving retrieval accuracy and motion quality compared with existing methods.

\subsection{Long-horizon Drift Analysis}
To directly evaluate long-horizon global drift, we generate a 5-min solo motion using the prompt ``a person walks straight forward.''
We compare ARMS with MotionStreamer~\cite{xiao2025motionstreamer} using two trajectory-level metrics.
Mean Lateral Deviation (MLD) measures the average absolute deviation from the expected forward direction, where lower is better.
Forward Efficiency (FE) measures the ratio between net forward progress and total traveled distance, where higher is better and the maximum value is 1.
\begin{table}[tb]
  \caption{Long-horizon drift analysis on 5-min solo motion generation.}
  \label{table:drift-analysis}
  \centering
  \begin{tabular}{lcc}
    \toprule
    Method & MLD (m) \(\downarrow\) & FE \(\uparrow\) \\
    \midrule
    MotionStreamer~\cite{xiao2025motionstreamer} & 22.39 & 0.363 \\
    Ours & \textbf{3.35} & \textbf{0.985} \\
    \bottomrule
  \end{tabular}
\end{table}
As shown in~\cref{table:drift-analysis}, ARMS substantially reduces lateral drift and maintains forward progress over the 5-min rollout.
Because interaction generation in ARMS builds on the anchor trajectory and relational displacement, stable solo anchor motion also supports stable anchor-based interaction rollout.

\subsection{Close-contact Penetration Analysis}
We further evaluate fine-grained interaction quality on a close-contact subset with contact frequency approximately 60\%.
We report average Penetration Depth (PD) and Penetration Frame Rate (PFR).
PD measures the average inter-body penetration depth in centimeters, while PFR measures the percentage of frames with penetration; lower values are better for both metrics.
\begin{table}[tb]
  \caption{Inter-body penetration analysis on the close-contact subset.}
  \label{table:close-contact}
  \centering
  \begin{tabular}{lcc}
    \toprule
    Method & PD (cm) \(\downarrow\) & PFR (\%) \(\downarrow\) \\
    \midrule
    Ground truth & 4.65$^{\pm0.00}$ & 18.35$^{\pm0.00}$ \\
    InterGen~\cite{liang2024intergen} & \underline{2.83$^{\pm0.09}$} & 28.25$^{\pm1.39}$ \\
    InterMask~\cite{javedintermask} & 3.04$^{\pm0.12}$ & \textbf{19.96$^{\pm0.80}$} \\
    Ours & \textbf{2.72$^{\pm0.05}$} & \underline{23.9$^{\pm1.02}$} \\
    \bottomrule
  \end{tabular}
\end{table}
As shown in~\cref{table:close-contact}, ARMS achieves the lowest penetration depth and lower penetration frame rate than InterGen, while InterMask obtains a slightly lower penetration frame rate.
Because contact precision is affected by the body model, body shape, mesh resolution, and the use of explicit physics constraints, we report these results as reference evidence for close-contact interaction quality.
ARMS does not include a physics-aware or contact-refinement module, so dedicated contact refinement remains a complementary direction for future work.

\subsection{Inference Efficiency}
\begin{table}[tb]
  \caption{Inference speed comparison.
  We report first-frame latency and the time required to generate a 10-second motion sequence for both solo and interaction settings.}
  \label{table:inference_time}
  \centering
    \resizebox{\linewidth}{!}{
   \begin{tabular}{l c c c c c }
    \toprule
    \multirow{2.5}{*}{Methods} &
    \multicolumn{2}{c}{\makecell{First frame latency \\ (second)}} & 
    \multicolumn{2}{c}{\makecell{10-second seq time \\ (second)}} &
    \multirow{2.5}{*}{Transition} \\
    \cmidrule{2-3} \cmidrule{4-5} 
    & Solo & Interaction & Solo & Interaction & \\
    \midrule
    MotionStreamer~\cite{xiao2025motionstreamer} &
    0.059$^{\pm.002}$&
    --&
    4.215$^{\pm.493}$&
    --&
    0.060$^{\pm.002}$\\
    InterMask~\cite{javedintermask} &
    --&
    0.339$^{\pm.011}$&
    --&
    0.339$^{\pm.011}$&
    --\\
    TIMotion~\cite{wang2025timotion} &
    -- &
    0.310$^{\pm.011}$&
    -- &
    0.310$^{\pm.011}$&
    -- \\
    \midrule
    Ours (non-streaming, $K=50$) &
    0.164$^{\pm.000}$&
    0.167$^{\pm.001}$&
    0.166$^{\pm.000}$&
    0.168$^{\pm.001}$&
    --
    \\
    Ours (streaming, $S=5, K=20$) &
    0.067$^{\pm.000}$&
    0.068$^{\pm.000}$&
    0.297$^{\pm.001}$&
    0.299$^{\pm.001}$&
    0.068$^{\pm.000}$\\
    Ours (streaming, $S=5, K=50$) &
    0.165$^{\pm.000}$&
    0.166$^{\pm.001}$&
    0.395$^{\pm.001}$&
    0.390$^{\pm.001}$&
    0.168$^{\pm.001}$\\
    Ours (streaming, $S=10, K=50$) &
    0.163$^{\pm.001}$&
    0.165$^{\pm.001}$&
    0.277$^{\pm.001}$&
    0.280$^{\pm.001}$&
    0.165$^{\pm.001}$\\
    \bottomrule
  \end{tabular}
  }
\end{table}

We evaluate the inference efficiency of our framework and compare it with representative streaming and interaction motion generation methods.
Following prior work~\cite{xiao2025motionstreamer}, we report the latency required to generate the first frame and the total time required to generate a 10-second motion sequence.
For streaming models, the first-frame latency measures the time needed to produce the first output frame after receiving a prompt, whereas for full-window models it corresponds to the time required to generate the entire sequence.

As shown in~\cref{table:inference_time}, MotionStreamer~\cite{xiao2025motionstreamer} achieves the lowest first-frame latency for single-person motion generation but requires substantially longer time to generate long motion sequences.
In contrast, our model with autoregressive streaming has a slightly higher first-frame latency, but generates long motions substantially faster due to the staggered noise scheduling strategy that allows later segments to be refined while earlier ones converge.
For interaction generation, InterMask~\cite{javedintermask} and TIMotion~\cite{wang2025timotion} generate the entire motion sequence in a single pass, leading to higher overall latency.
Our non-streaming configuration (denoted as full-window generation) generates the entire sequence within a single denoising window, similar to InterMask and TIMotion, and therefore exhibits nearly constant generation time independent of sequence length.
Notably, it is faster than both InterMask and TIMotion for interaction generation.
In contrast, the streaming configurations generate motion incrementally using a segment-based schedule, and the inference time therefore scales with the sequence length.
With a reduced number of sampling steps ($K=20$), the model achieves a lower first-frame latency (0.067\,s) and faster overall sequence generation.
Larger segments (e.g., $S=10$) allow more frames to be refined jointly during each denoising step, further reducing the total generation time for long sequences compared with smaller segments.
Importantly, our framework supports transitions between solo motion and interaction within the same generation process.
The transition latency reported in~\cref{table:inference_time} shows that the model can switch between agent configurations with minimal overhead, enabling efficient streaming generation in dynamic social scenarios.

\section{Qualitative Results}
In addition to the quantitative evaluations, we provide further qualitative demonstrations on the project website.
Since motion is inherently temporal, static images cannot fully capture the dynamics of the generated sequences.
We therefore present the qualitative results in video format.
The videos illustrate streaming generation across various scenarios, including incremental generation of solo motions, human--human interactions, transitions between solo and interaction behaviors, and comparisons with other methods.

\section{Representation}
Different interaction motion generation methods adopt distinct motion representations, which influence spatial consistency and interaction quality.
To clarify these differences, we summarize the design of representative representations in~\cref{tab:rep} and provide qualitative comparisons on the project website.
\begin{table}[t]
\centering
\caption{Comparison of interaction motion representations.}
\label{tab:rep}
\resizebox{\linewidth}{!}{
\begin{tabular}{l c c c}
\toprule
Method & Per-agent motion formulation & Interaction modeling & Observed behavior \\
\midrule
Non-canonical~\cite{liang2024intergen} 
& $
  \mathbf{x} =
  \left[
    \mathbf{p}^{global},
    \dot{\mathbf{p}}^{global},
    \mathbf{q},
    \mathbf{c}
  \right]
  $ 
& Implicit in global coordinates 
& \makecell{Strong spatial coupling but restricted to coordinate-dependent \\ patterns and prone to foot sliding} \\
\midrule
Canonical~\cite{shafirhuman, yusocialgen}
& $ 
  \mathbf{x} =
  \left[
    \mathbf{v}^{yaw},
    \mathbf{v}^{xz},
    \mathbf{p}^{local},
    \dot{\mathbf{p}}^{local},
    \mathbf{q},
    \mathbf{c}
  \right]
  $
& \makecell{Initial relative transform \\ between agents ($\Delta^{xz}_0, \Delta^{yaw}_0$)}
& \makecell{Smooth individual motion \\ but weak interaction consistency} \\
\midrule
Ours 
& $
  \mathbf{x} =
  \left[
    \mathbf{r}^{yaw},
    \mathbf{v}^{xz},
    \mathbf{p}^{local},
    \dot{\mathbf{p}}^{local},
    \mathbf{q},
    \mathbf{c}
  \right]
  $
& \makecell{Per-frame relative displacement \\ between agents ($\Delta^{xz}$)}
& \makecell{Stable interaction and temporally extendable \\ generation, but with minor foot sliding} \\
\bottomrule
\end{tabular}
}

\vspace{2pt}
\parbox{\linewidth}{%
{\scriptsize
\textbf{Notation:}
$\mathbf{p}$ denotes joint positions.
$\dot{\mathbf{p}}$ denotes joint velocities.
$\mathbf{q}$ represents joint rotations.
$\mathbf{v}$ denotes root velocities, including yaw angular velocity $\mathbf{v}^{yaw}$ and planar velocity $\mathbf{v}^{xz}$.
$\mathbf{r}^{yaw}$ denotes root yaw rotation.
$\mathbf{c}$ denotes foot contact indicators.
$\Delta^{xz}$ represents planar relative displacement between agents.
}%
}
\end{table}


\end{document}